\documentclass{article}

\PassOptionsToPackage{numbers, compress}{natbib}
\usepackage[preprint]{neurips_2026}
\makeatletter
\renewcommand{\@noticestring}{Under review.}
\makeatother

\usepackage[utf8]{inputenc} 
\usepackage[T1]{fontenc}    

\usepackage{hyperref}       
\usepackage{url}            
\usepackage{booktabs}       
\usepackage{float}          
\usepackage{algorithm}      
\usepackage{algpseudocode} 
\usepackage{enumitem}       
\usepackage{array}          
\usepackage{multirow}       
\usepackage{amsmath}        
\usepackage{amsfonts}       
\usepackage{bm}             
\usepackage{nicefrac}       
\usepackage{microtype}      
\usepackage{xcolor}         
\usepackage{tikz}
\usepackage{xspace}
\usepackage{pifont}         
\newcommand{\cmark}{\textcolor{green!60!black}{\ding{51}}}
\newcommand{\pmark}{\textcolor{gray}{\textbf{?}}}
\newcommand{\xmark}{\textcolor{red!70!black}{\ding{55}}}

\usepackage{graphicx}
\usepackage{subcaption} 
\graphicspath{{./figs/}}
\usepackage{placeins} 
\usepackage{wrapfig}  

\title{HypergraphFormer: Learning Hypergraphs from LLMs for Editable Floor Plan Generation}

\author{%
  Nikita Klimenko\textsuperscript{*} \\
  Autodesk Research\\
  \texttt{nikita.klimenko@autodesk.com} \\
  \And
  Hesam Salehipour\textsuperscript{*,\dag} \\
  Autodesk Research\\
  \texttt{hesam.salehipour@autodesk.com} \\
  \AND
  Parham Eftekhar\textsuperscript{\ddag} \\
  York University \\
  \texttt{eftekhar@yorku.ca} \\
  \And
  Amir Khasahmadi \\
  Autodesk Research\\
  \texttt{amir.khasahmadi@autodesk.com} \\
  \And
  Ramon Elias Weber \\
  UC Berkeley \\
  \texttt{ramon@berkeley.edu} \\
}

\definecolor{RoomLiving}{HTML}{ED6D87}      
\definecolor{RoomKitchen}{HTML}{9A5387}     
\definecolor{RoomBedroom}{HTML}{5B448C}     
\definecolor{RoomBathroom}{HTML}{FFE953}    
\definecolor{RoomEntrance}{HTML}{ED6D87}    
\definecolor{RoomStorage}{HTML}{FF9FBB}     
\definecolor{RoomDoor}{HTML}{B8B5B4}        

\DeclareRobustCommand{\bliving}{\tikz[baseline=-0.05ex]\draw[fill=RoomLiving,draw=black,line width=0.3pt] (0,0) rectangle (0.8em,0.8em);\xspace}
\DeclareRobustCommand{\bkitchen}{\tikz[baseline=-0.05ex]\draw[fill=RoomKitchen,draw=black,line width=0.3pt] (0,0) rectangle (0.8em,0.8em);\xspace}
\DeclareRobustCommand{\bbedroom}{\tikz[baseline=-0.05ex]\draw[fill=RoomBedroom,draw=black,line width=0.3pt] (0,0) rectangle (0.8em,0.8em);\xspace}
\DeclareRobustCommand{\bbathroom}{\tikz[baseline=-0.05ex]\draw[fill=RoomBathroom,draw=black,line width=0.3pt] (0,0) rectangle (0.8em,0.8em);\xspace}
\DeclareRobustCommand{\bentrance}{\tikz[baseline=-0.05ex]\draw[fill=RoomEntrance,draw=black,line width=0.3pt] (0,0) rectangle (0.8em,0.8em);\xspace}
\DeclareRobustCommand{\bstorage}{\tikz[baseline=-0.05ex]\draw[fill=RoomStorage,draw=black,line width=0.3pt] (0,0) rectangle (0.8em,0.8em);\xspace}
\DeclareRobustCommand{\bdoor}{\tikz[baseline=-0.05ex]\draw[fill=RoomDoor,draw=black,line width=0.3pt] (0,0) rectangle (0.8em,0.8em);\xspace}

\begin{document}

\maketitle
\renewcommand{\thefootnote}{\fnsymbol{footnote}}
\footnotetext[1]{Equal contribution.}
\footnotetext[2]{Corresponding author.}
\footnotetext[3]{This study was conducted during the author's internship at Autodesk.}
\renewcommand{\thefootnote}{\arabic{footnote}}
\setcounter{footnote}{0}


\begin{abstract}
    In this work, we propose \emph{HypergraphFormer}, a novel and efficient approach to floor plan generation based on learning hypergraph representations with a large language model (LLM). The model is trained via supervised fine-tuning to generate a hypergraph-based textual representation that encodes spatial relationships and connectivity information within floor plans.
    We train and evaluate our approach on the RPLAN dataset, and further demonstrate its generalizability on a separate out-of-distribution dataset, which we release in this paper.
    Our method outperforms state-of-the-art techniques based on rasterized or vectorized representations across a diverse set of metrics. We also show improved data efficiency, particularly under distribution shift.
    The hypergraph formulation enables the generation of floor plans for arbitrary, irregular, user-specified boundaries by decoupling apartment footprints from their functional and geometric subdivisions.
    Furthermore, we show that the proposed methodology offers a high degree of editability, making it particularly well suited to design-oriented workflows supported by LLMs.
\end{abstract}

\section{Introduction}
\label{sec:intro}

The design of architectural floor plans is challenging and time-consuming, requiring a manual, iterative process in which professional architects balance competing requirements~\cite{heckmann2017floor}. Given the unprecedented scale of global urbanization, automated methods for floor plan analysis and generation are a key opportunity to make architectural design more accessible and to produce higher-quality indoor spaces at scale. Existing deep-learning approaches to floorplan generation commit to one of two output representations, both with structural drawbacks: raster-based methods generate floor plans as low-resolution pixel masks~\cite{housegan,housegan++,Tang2023GraphTransformer,MaskPlan,iplan,hu2020graph2plan}, where the spatial resolution is bounded by the pixel size, walls and openings must be encoded implicitly, and the output requires non-trivial post-processing to recover a usable plan; vector-based methods predict per-room polygons or coordinate pairs~\cite{housediffusion,DiffPlanner,wallplan,GSDiff,QIU2025,housetune}, but typically restrict the layout to axis-aligned rectangles inside a fixed boundary and still depend on geometric post-processing for closure and consistency. Beyond representation, a second axis matters in practice: the cost of \emph{editing} a generated plan and the ability to \emph{generalize} beyond the training distribution. As we summarize in Table~\ref{tab:related_work_comparison}, every prior baseline either lacks fine-grained or boundary-level editability or recovers it only through dedicated training-time machinery and a full re-inference pass, none natively expresses arbitrary non-axis-aligned wall geometry, and none reports data-efficient training or out-of-distribution evaluation. Recent work has begun to leverage Large Language Models (LLMs) for layout synthesis~\cite{housetune,tell2design}, but existing LLM-based pipelines hand the geometry off to a diffusion or raster decoder and inherit the same limitations. See Appendix~\ref{sec:related} for a detailed discussion of related work.

\begin{table*}[t]
    \centering
    \scriptsize
    \renewcommand{\arraystretch}{1.15}
    \setlength{\tabcolsep}{0.7pt}
    \caption{Feature comparison of representative floor plan generation methods. ``Boundary'' refers to the building outline (plus entrance/front-door when supplied). \textbf{RC}~=~room counts, \textbf{RT}~=~room types/labels, \textbf{RS}~=~room sizes/areas, \textbf{RL}~=~room locations/centers. For input/output columns, \cmark{}/\xmark{} denote whether the pathway is supported. For editability and extra-features columns, \cmark{}/\pmark{}/\xmark{} denote full, partial, and no support, respectively, as documented in the cited papers.}
    \label{tab:related_work_comparison}
    \begin{tabular}{l c c c l c c c c c c c c c}
        \toprule
        & \multicolumn{4}{c}{\textbf{Input}}
        & \multicolumn{3}{c}{\textbf{Output}}
        & \multicolumn{3}{c}{\textbf{Editability}}
        & \multicolumn{3}{c}{\textbf{Extra Features}} \\
        \cmidrule(lr){2-5} \cmidrule(lr){6-8} \cmidrule(lr){9-11} \cmidrule(lr){12-14}
        Paper
        & \rotatebox[origin=l]{60}{\shortstack[l]{Access\\graph}}
        & \rotatebox[origin=l]{60}{\shortstack[l]{Adjacency\\graph}}
        & \rotatebox[origin=l]{60}{Boundary}
        & \rotatebox[origin=l]{60}{Other}
        & \rotatebox[origin=l]{60}{Raster}
        & \rotatebox[origin=l]{60}{Vector}
        & \rotatebox[origin=l]{60}{Graph}
        & \rotatebox[origin=l]{60}{\shortstack[l]{Boundary\\Editability}}
        & \rotatebox[origin=l]{60}{\shortstack[l]{Fine-grained\\Editability}}
        & \rotatebox[origin=l]{60}{\shortstack[l]{Irregular\\Boundaries}}
        & \rotatebox[origin=l]{60}{\shortstack[l]{Data\\Efficiency}}
        & \rotatebox[origin=l]{60}{\shortstack[l]{Out-of-\\Distribution}} \\
        \midrule
        House-GAN~\cite{housegan}
            & \xmark & \cmark & \xmark & --
            & \cmark & \xmark & \xmark
            & \xmark & \xmark & \xmark
            & \xmark & \xmark \\
        House-GAN++~\cite{housegan++}
            & \cmark & \xmark & \xmark & --
            & \cmark & \xmark & \xmark
            & \xmark & \xmark & \xmark
            & \xmark & \xmark \\
        Graph Transformer GANs~\cite{Tang2023GraphTransformer}
            & \xmark & \cmark & \xmark & --
            & \cmark & \xmark & \xmark
            & \xmark & \xmark & \xmark
            & \xmark & \xmark \\
        MaskPLAN~\cite{MaskPlan}
            & \xmark & \cmark & \cmark & {\tiny RT, RL, RS}
            & \cmark & \xmark & \xmark
            & \pmark & \pmark & \xmark
            & \xmark & \xmark \\
        iPLAN~\cite{iplan}
            & \xmark & \xmark & \cmark & {\tiny RT, RC, RL, RS}
            & \cmark & \xmark & \xmark
            & \pmark & \pmark & \pmark
            & \xmark & \xmark \\
        WallPlan~\cite{wallplan}
            & \xmark & \cmark & \cmark & --
            & \xmark & \cmark & \xmark
            & \pmark & \xmark & \xmark
            & \xmark & \xmark \\
        HouseDiffusion~\cite{housediffusion}
            & \cmark & \xmark & \xmark & --
            & \xmark & \cmark & \xmark
            & \xmark & \xmark & \pmark
            & \xmark & \xmark \\
        HouseTune~\cite{housetune}
            & \xmark & \xmark & \xmark & {\tiny RT, RL, RS}
            & \xmark & \cmark & \xmark
            & \xmark & \xmark & \xmark
            & \xmark & \xmark \\
        DiffPlanner~\cite{DiffPlanner}
            & \xmark & \cmark & \cmark & {\tiny RC, RT, RS, RL}
            & \xmark & \cmark & \xmark
            & \pmark & \pmark & \xmark
            & \xmark & \xmark \\
        GSDiff~\cite{GSDiff}
            & \xmark & \cmark & \cmark & --
            & \xmark & \cmark & \xmark
            & \pmark & \xmark & \pmark
            & \xmark & \xmark \\
        Graph2Plan~\cite{hu2020graph2plan}
            & \xmark & \cmark & \cmark & {\tiny RC, RT, RL}
            & \cmark & \xmark & \xmark
            & \pmark & \pmark & \xmark
            & \xmark & \xmark \\
        \midrule
        \textbf{HypergraphFormer (Ours)}
            & \cmark & \xmark & \xmark & --
            & \xmark & \xmark & \cmark
            & \cmark & \cmark & \cmark
            & \cmark & \cmark \\
        \bottomrule
    \end{tabular}
\end{table*}

This paper takes an empirical stance: rather than proposing a new architecture or a new floor plan representation, we ask what becomes possible when a small LLM is trained to generate an \emph{existing} structured representation. We adopt the graph-based textual representation of a floor plan introduced by \citet{hypergraph}, referred to as a \emph{hypergraph}, which decouples an apartment's outer boundary from its interior layout by combining a binary space partition (BSP) tree, hierarchically representing the spatial decomposition into rooms, with an access graph capturing their functional connectivity. We show that a small instruction-tuned LLM can be fine-tuned to produce this representation directly, which unlocks a combination of properties that prior raster- and vector-based pipelines lack: native editability of the generated plan, strong out-of-distribution generalization to architect-designed apartments, and substantial data efficiency relative to state-of-the-art baselines. We refer to the resulting system as \emph{HypergraphFormer}, and our contributions are:
\begin{itemize}
    \item \textbf{Representation.} We show that a lightweight LLM can be fine-tuned to generate the structured hypergraph representation of \citet{hypergraph} directly, yielding a generative pipeline that produces floor plans conforming to arbitrary, user-specified boundaries without rasterization or per-room polygon prediction.
    \item \textbf{Editability.} We show that this representation enables a family of fast, LLM-free procedural edits (e.g. adding or removing rooms, rotation/reflection, and gradient-descent refinement of room areas) that compose with the generated plan and further improve layout quality, as well as higher-level edits applied via LLM tool calls on the BSP tree and access graph.
    \item \textbf{Data efficiency.} We conduct a controlled data-efficiency study showing that our approach matches the accuracy of state-of-the-art baselines using only a small fraction of their training data.
    \item \textbf{Out-of-distribution generalization.} We demonstrate strong out-of-distribution generalization: trained on RPLAN, our model exceeds these baselines on \emph{WMR24}, our curated dataset of architect-designed floor plans whose distribution of apartment size, bedroom count, and boundary geometry differs substantially from the training distribution. We release WMR24 together with our training, inference, dataset-conversion, and procedural-editing code to enable full reproduction and extension of these results.
\end{itemize}

\section{Methodology}
\label{sec:method}
As introduced by \citet{hypergraph}, a hypergraph is a reduced-order representation for floor plans: each apartment is decomposed into a boundary and a hypergraph stored as structured JSON, in which intermediate nodes encode binary space partitions of the interior and leaf nodes carry per-room semantics together with an access graph over door connections. We adopt this representation throughout and refer the reader to Appendix~\ref{sec:hypergraph} for full details.

Our core methodological contribution is to bring this compact, structured hypergraph representation into a learning-based generation pipeline powered by instruction-tuned LLMs. By predicting the hypergraph (rather than pixels or raw geometry), we obtain outputs that are easier to validate and edit, and more compatible with downstream procedural reconstruction, while also improving quantitative performance relative to common image-based baselines.

\begin{figure*}[t]
    \centering
    \subcaptionbox{}{%
        \includegraphics[height=4.5cm,keepaspectratio]{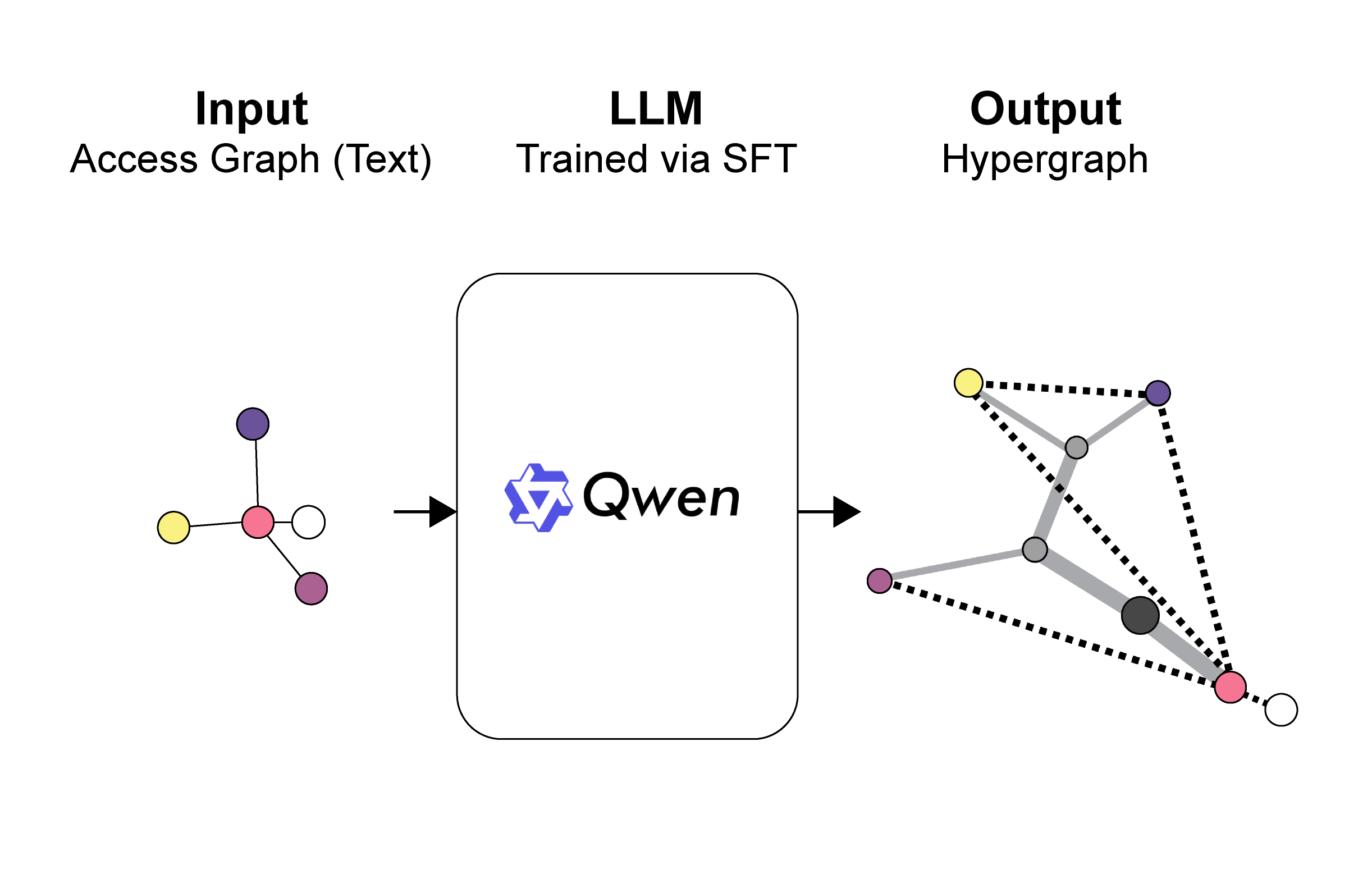}%
    }\hspace{1em}%
    \subcaptionbox{}{%
        \includegraphics[height=4.5cm,keepaspectratio]{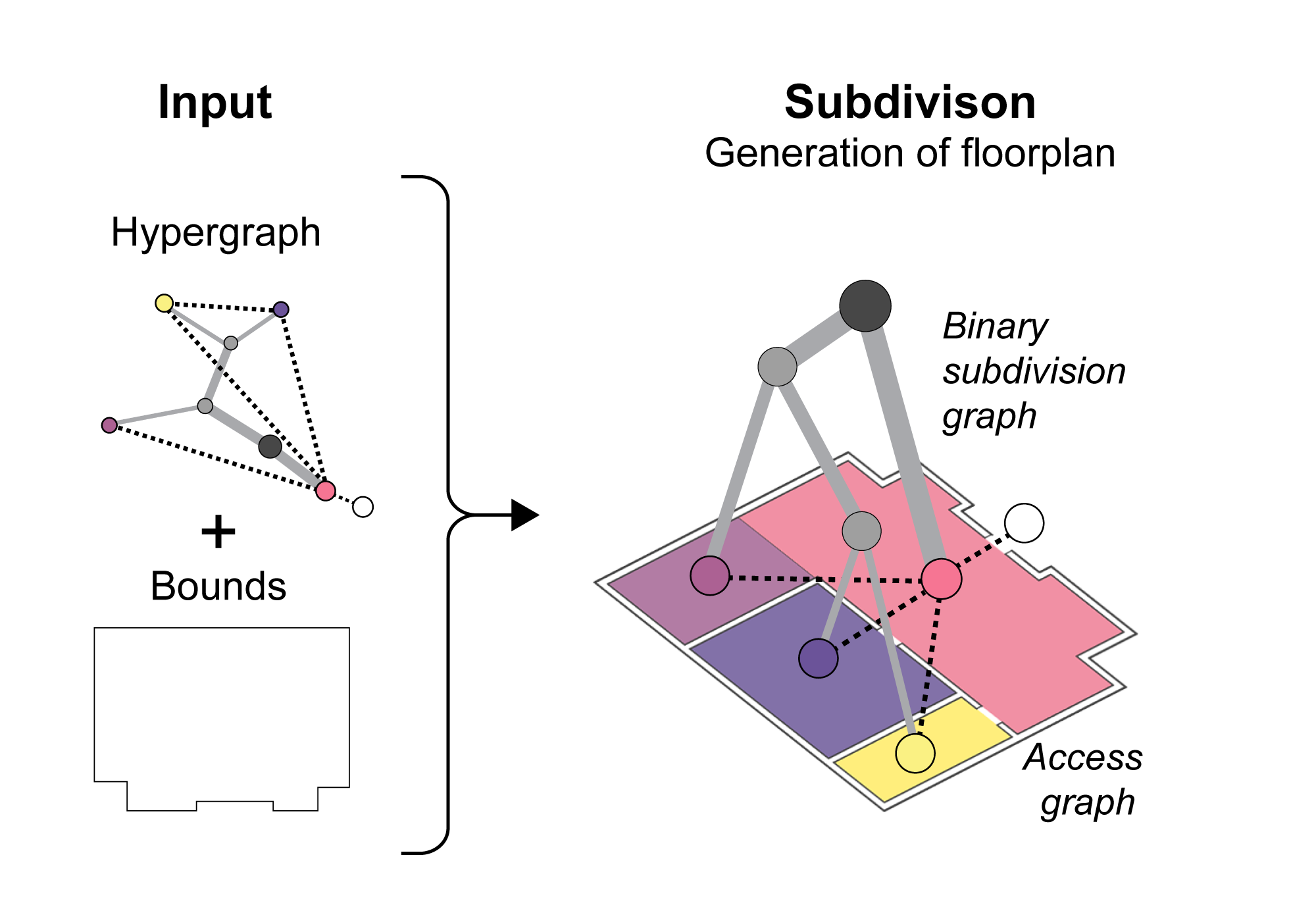}%
    }
    \caption{An overview of HypergraphFormer for floor-plan generation. (a) Supervised fine-tuning on (access graph, hypergraph) pairs. (b) Generation of floor-plan from hypergraph and bounds at inference time.}
    \label{fig:HypergraphFormer}
\end{figure*}

\subsection{Supervised fine-tuning} 
\label{sec:SFT}
%
We perform supervised fine-tuning to map an access graph to a hypergraph that satisfies the representation constraints. We formulate this as constrained structured generation: each prompt explicitly specifies the required JSON fields together with the structural invariants of the representation (BSP tree, leaf-only connectivity, area conservation, and axis-aligned splits), followed by an instruction to generate a hypergraph for a given access graph, with the corresponding ground-truth hypergraph as the target output. As depicted in Fig.~\ref{fig:HypergraphFormer}, we fine-tune an open-source LLM, Qwen3~\cite{yang2025qwen3}, with LoRA to produce a hypergraph from the input access graph. This stage teaches the model the underlying structural representation required to produce editable floor plans and provides a foundation for the procedural edits that further refine the generated structures with respect to room compactness, area allocation, and room connectivity. See Appendix~\ref{app:implementation_details} for further details.

We fine-tune the model on the RPLAN dataset \cite{RPLAN}, using the raw splits provided by~\cite{DiffPlanner} for training, validation, and testing; Appendix~\ref{app:rplan_to_hypergraph} describes how we convert the raw RPLAN samples into our hypergraph format. For out-of-distribution evaluation, we additionally test on \emph{WMR24} \cite{hypergraph}, a curated dataset of architect-designed floor plans whose boundary shapes and design conventions differ markedly from RPLAN; see Appendix~\ref{app:dataset_wmr24} for details on the dataset and its construction.

\subsection{Evaluation metrics}
\label{sec:metrics}
For each test apartment $i = 1, \dots, N$, let $D_i$ denote the ground-truth apartment, $P_i$ the prediction, and $B_i$ the apartment boundary polygon. Below, we drop the apartment index $i$ for brevity. We also define $|a|$ for a polygon's area and $\pi(a) = |a|/|B|$ for its share of the apartment. Let $\mathcal{T}$ be the set of room types and $D_{t}, P_{t}$ the type-$t$ sub-multisets of $D$ and $P$. To compare per-room geometry, we form a matched-pair set $\mathcal{M} = \bigcup_{t \in \mathcal{T}} \mathcal{M}_{t}$, where $\mathcal{M}_{t}$ pairs $P_{t}$ with $D_{t}$ by sorting both in descending order of area and matching index-wise; we additionally write $\mathcal{M}^{\star}$ for the same construction with the shorter side padded by zero-area phantoms, so that any missing or surplus room contributes its full proportion. Our metrics fall into two groups: \emph{structural} metrics on the access graph and \emph{geometric/tiling} metrics on the predicted polygons. We especially emphasize the use of structural metrics: for floor-plan generation tasks that are based on input room connectivity rules (hereafter referred to as access graphs) or an input set of rooms, checking adherence to these rules is imperative. The geometric metrics complement the structural metrics by describing the realism of the generated rooms in terms of sufficient area and geometric properties.

\paragraph{Structural metrics.}
Let $g(A)$ denote the access graph extracted from an apartment $A$. We measure how faithfully a prediction reproduces the input access graph using the standard graph edit distance~\cite{GED},
\begin{equation}
\mathrm{GED}(g_1, g_2)
= \min_{(o_1,\ldots,o_k)\in \mathcal{S}_{\text{label}}(g_1,g_2)}
\sum_{j=1}^{k} c(o_j),
\label{eq:ged}
\end{equation}
the minimum-cost edit sequence (vertex/edge insertion, deletion, label-preserving substitution) that transforms $g_1$ into $g_2$, with $c(o_j)$ a per-operation cost. From it we derive two test-set accuracies, the strict access-graph accuracy $\mathcal{A}$ and a type-and-count multiset accuracy $\mathcal{A}_{tc}$ that relaxes the connectivity check:
\begin{equation}
    \mathcal{A}
    = \tfrac{1}{N}\!\sum_{i=1}^{N}\! \mathbf{1}\!\big[\,\mathrm{GED}(g(P_i), g(D_i)) = 0\,\big],
    \qquad
    \mathcal{A}_{tc}
    = \tfrac{1}{N}\!\sum_{i=1}^{N}\! \mathbf{1}\!\big[\,|P_{i,t}| = |D_{i,t}|\;\forall\, t \in \mathcal{T}\,\big].
    \label{eq:accuracy}
\end{equation}
By construction $\mathcal{A} \leq \mathcal{A}_{tc}$, since matching the access graph implies matching per-type counts. We treat $\mathcal{A}$ as the strictest accuracy in the paper.

\paragraph{Geometric and tiling metrics.}
We adopt the per-polygon compactness deviation $\delta(a, b) = 1 - L_{Sa} L_b / L_a L_{Sb}$ of~\cite{hypergraph}, where $L_a, L_b$ are the perimeters of polygons $a, b$ and $L_{Sa}, L_{Sb}$ the perimeters of squares with the same areas as $a$ and $b$, respectively, and complement it with a scale-invariant area-proportion error $\varepsilon$. Aggregated over the matched-pair sets defined above,
\begin{align}
\delta(P, D) &= \frac{1}{|\mathcal{M}|} \sum_{(p, d) \in \mathcal{M}} \delta(p, d), \label{eq:delta} \\
\varepsilon(P, D) &= \frac{1}{|\mathcal{M}^{\star}|} \sum_{(p, d) \in \mathcal{M}^{\star}} \big|\pi(p) - \pi(d)\big|, \label{eq:epsilon}
\end{align}
where $\delta$ compares only matched pairs (it is a shape statistic, so missing or surplus rooms have no counterpart to compare against), while $\varepsilon$ uses the phantom-padded set so that missing or surplus rooms incur their full area proportion as error. Finally, to verify that the predicted rooms cover the input boundary $B$ without spilling out or overlapping, we report two boundary-normalized tiling ratios,
\begin{equation}
\rho_{\mathrm{out}}(P, B) = \frac{|U \setminus B|}{|B|},
\qquad
\rho_{\mathrm{ovl}}(P, B) = \frac{S - |U|}{|B|},
\label{eq:tiling}
\end{equation}
where $U = \bigcup_{p \in P} p$ is the union of predicted room polygons and $S = \sum_{p \in P} |p|$ is the sum of their individual areas; $\rho_{\mathrm{out}}$ measures predicted area spilling outside $B$ and $\rho_{\mathrm{ovl}}$ measures inter-room overlap. All four metrics are non-negative and reported as test-set means; $\varepsilon, \rho_{\mathrm{out}}, \rho_{\mathrm{ovl}}$ are given as percentages. We have $\rho_{\mathrm{out}} = \rho_{\mathrm{ovl}} = 0$ iff the predicted rooms form an exact non-overlapping cover of $B$, and $\varepsilon = 0$ iff the predicted and ground-truth apartments have identical type-wise area distributions.

We do not report the Fr\'echet Inception Distance (FID), which prior raster floor-plan work~\cite{housegan++,housediffusion} adopts as a ``diversity'' score; the structural and geometric metrics above are more directly diagnostic of floor-plan validity, and we discuss this choice in detail in Appendix~\ref{app:fid_inadequacy}.

\subsection{Procedural editing}
\label{sec:procedural-editing}
%
Our hypergraph representation enables a family of lightweight editing operations that can be applied procedurally to floor plans generated by HypergraphFormer, without the need to train the LLM again. We use three such operations as a post-processing pipeline: (i) adding or removing a room to align the predicted multiset with the input access graph, (ii) rotating or flipping the hypergraph to maximize per-room compactness, and (iii) parametric optimization of the BSP split parameters by gradient descent. Full descriptions and pseudocode are given in Appendix~\ref{app:editing_tasks}.

\section{Results}
\label{sec:results}

We compare our method against four recent baselines: \emph{boundary-free} methods (HouseGAN++ (HG)~\cite{housegan++}, HouseDiffusion (HD)~\cite{housediffusion}), which take an access graph and produce a free-form layout; and \emph{boundary-constrained} methods (iPLAN (IP)~\cite{iplan}, DiffPlanner (DP)~\cite{DiffPlanner}), which take a boundary polygon and a room set and fill that boundary (cf.\ Table~\ref{tab:related_work_comparison}). Even though the latter methods also allow to input a room graph, they are trained on an adjacency graph rather than a connectivity graph, which prevents us from using it as input for a fair comparison. HypergraphFormer is evaluated against both of these groups: it consumes the same access graph as the first group and produces a hypergraph instantiable against any boundary, while for the second group we first prompt the same fine-tuned LLM to convert the input room set into an access graph. To keep each comparison fair, the metric subsets differ. Against boundary-free baselines we report only the structural metrics, GED~\eqref{eq:ged}, the access graph accuracy $\mathcal{A}$, and the room-set multiset accuracy $\mathcal{A}_{tc}$~\eqref{eq:accuracy}. Against boundary-constrained baselines, which do not rely on an input access graph, we report $\mathcal{A}_{tc}$ together with the geometric metrics $\delta$~\eqref{eq:delta} and $\varepsilon$~\eqref{eq:epsilon}. All baselines use their official checkpoints, all models are evaluated under the same protocol, and the evaluation code is released alongside the paper. Table~\ref{tab:input_grouped_comparison} reports dataset-level metrics for the two baseline groups; per room-count breakdowns are deferred to Appendix~\ref{app:per_bin_results} (Tables~\ref{tab:per_bin_comparison} and~\ref{tab:per_bin_comparison_boundary}).

\begin{figure}[!htbp]
    \centering
    \includegraphics[width=\textwidth,keepaspectratio]{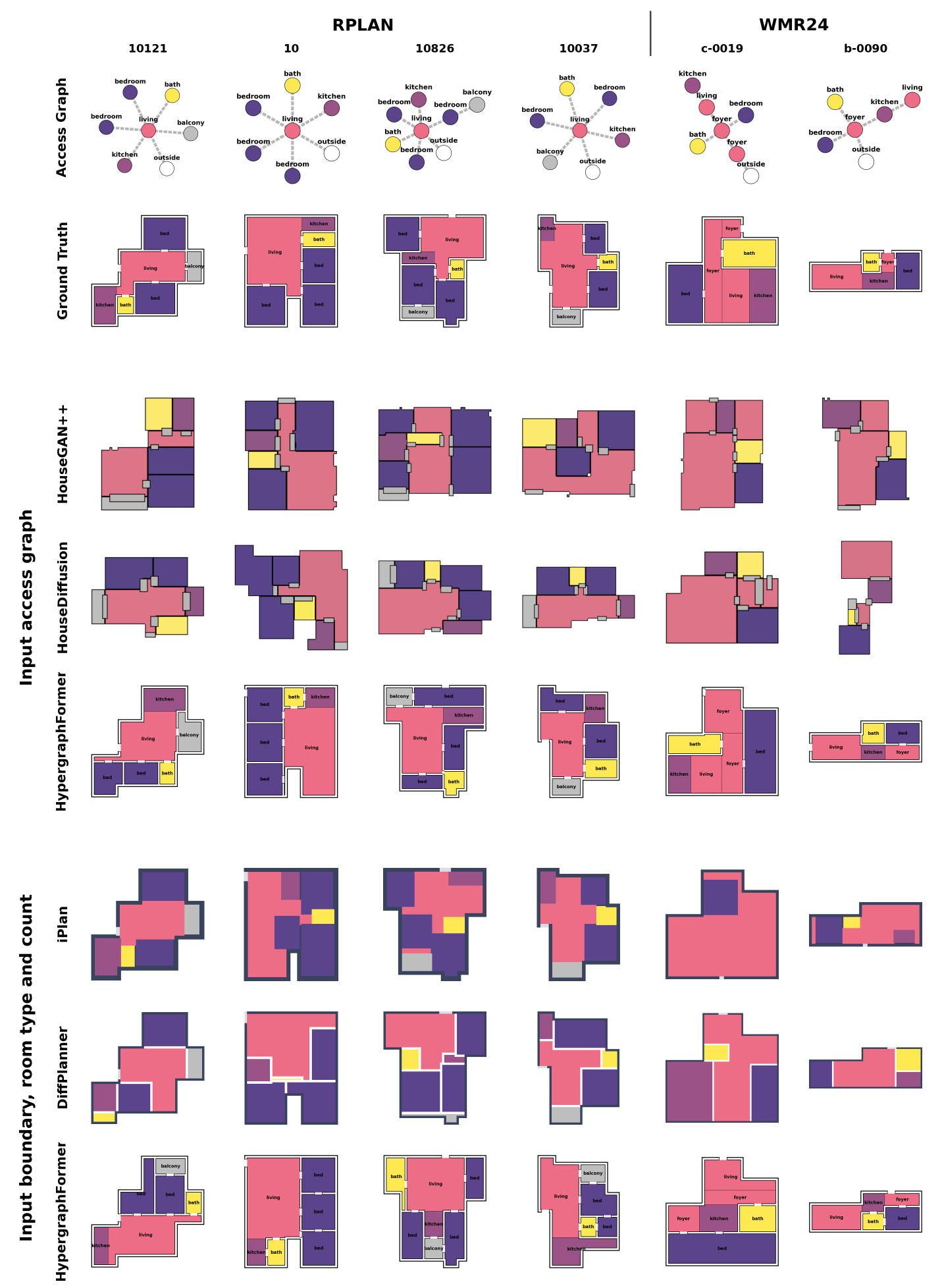}
    \caption{Qualitative comparison of generated floor plans. From left to right: (a) access graph, (b) ground truth, (c) HouseDiffusion, (d) HouseGAN++, (e) HypergraphFormer. Rooms are colored by their function, namely: living room \bliving~, kitchen \bkitchen~, bedroom \bbedroom~, bathroom \bbathroom~, entrance \bentrance~, storage \bstorage~, interior door \bdoor~.}
    \label{fig:visual_comparison}
\end{figure}

\paragraph{Visual comparison of predicted floor plans.} 

Fig.~\ref{fig:visual_comparison} presents a visual comparison of generated floor plans for several test-set samples, alongside predictions from our method. For each ground truth floorlpan, we demonstrate predictions for boundary-free and boundary-constrained methods with appropriate baselines. The first row shows the input access graphs, where nodes correspond to rooms (color-coded by program) and edges represent room connectivity. The remaining rows display the ground-truth apartment layouts and the corresponding predictions from each model, as indicated.

As illustrated, both HouseGAN++ and HouseDiffusion assign apartment footprints randomly, leading to noticeable inconsistencies across examples. In contrast, owing to its underlying hypergraph-based representation, our approach allows users to explicitly specify the outer boundary, and the generated hypergraph is then fitted to the prescribed space. Moreover, unlike HouseGAN++ and HouseDiffusion, which directly generate rasterized or vectorized images, HypergraphFormer outputs a structured representation that can be directly imported, edited, and modeled within off-the-shelf architectural design tools such as Rhino and Grasshopper. 

We also observe characteristic qualitative artifacts in the baseline methods. HouseGAN++ frequently produces noisy and overly complex space partitions, such as bedrooms splitting the living room (\textit{10121}) or bathrooms placed in the middle of the apartment (\textit{10826}). HouseDiffusion generates either overlapping rooms that visually obscure one another (e.g., the bathroom and bedroom in \textit{10037}) or excessively spread-out layouts that do not conform to typical apartment boundaries (\textit{b-0090}). In contrast, HypergraphFormer avoids both failure modes: rooms are derived from a binary space partition of the input boundary, ensuring that the predicted polygons tile the floor plan exactly, with no gaps or overlaps by construction, thereby yielding more visually plausible and well-distributed layouts.

Although boundary-constrained approaches do respect the apartment boundary, they rely on limited room-placement mechanisms, which often produce artifacts requiring substantial editing and reinterpretation. iPlan tends to place rooms in an ad hoc manner, frequently splitting living-room space into disjoint segments (as seen in samples \textit{10}, \textit{10826}, and \textit{b-0090}), while DiffPlanner often produces layouts with severe room overlaps (samples \textit{10} and \textit{10826}).

\begin{table*}[!htbp]
    \centering
    \footnotesize
    \caption{Dataset-level comparison on RPLAN and the out-of-distribution WMR24 test set, grouped by the inputs each method consumes. \textbf{Access graph} (top): boundary-free baselines HouseGAN++ (HG) and HouseDiffusion (HD)~\cite{housegan++,housediffusion}, evaluated on the structural metrics GED~\eqref{eq:ged}, GED accuracy $\mathcal{A}$, and joint type-and-count multiset accuracy $\mathcal{A}_{tc}$~\eqref{eq:accuracy}. \textbf{Boundary, RT, RC} (bottom): boundary-constrained baselines DiffPlanner~\cite{DiffPlanner} and iPLAN~\cite{iplan}, evaluated on $\mathcal{A}_{tc}$ together with the geometric and tiling metrics $\delta$~\eqref{eq:delta}, $\varepsilon$~\eqref{eq:epsilon}, and the boundary-normalized $\rho_{\mathrm{out}},\rho_{\mathrm{ovl}}$~\eqref{eq:tiling}. RT~=~room types, RC~=~room counts (cf.\ Table~\ref{tab:related_work_comparison}); $\mathcal{A}$, $\mathcal{A}_{tc}$, $\varepsilon$, $\rho_{\mathrm{out}}$, $\rho_{\mathrm{ovl}}$ are reported in percent.}
    \renewcommand{\arraystretch}{1.1}
    \setlength{\tabcolsep}{5pt}
    \begin{tabular}{@{}l l c c c c c c@{}}
        \toprule
        Model Inputs & Metric
            & \multicolumn{3}{c}{\textbf{RPLAN}}
            & \multicolumn{3}{c@{}}{\textbf{Out of Distribution (WMR24)}} \\
        \cmidrule(lr){3-5} \cmidrule(l){6-8}
        \multirow{4}{*}{Access graph}
            &                                          & HouseGAN++ & HouseDiffusion & Ours    & HouseGAN++ & HouseDiffusion & Ours   \\
            \cmidrule(lr){3-5} \cmidrule(l){6-8}
            & GED ($\downarrow$)                       & $2.59$ & $1.95$ & $\bm{1.62}$  & $3.80$ & $3.78$ & $\bm{1.70}$ \\
            & $\mathcal{A}$ (\%, $\uparrow$)           & $6.0$  & $16.3$ & $\bm{40.9}$  & $8.5$  & $2.6$  & $\bm{52.5}$ \\
            & $\mathcal{A}_{tc}$ (\%, $\uparrow$)      & $44.2$ & $96.7$ & $\bm{100.0}$ & $37.1$ & $80.0$ & $\bm{99.9}$ \\
        \midrule
        \midrule
        \multirow{6}{*}{\shortstack[l]{Boundary,\\ RT, RC}}
            &                                          & iPLAN   & DiffPlanner & Ours    & iPLAN & DiffPlanner & Ours   \\
            \cmidrule(lr){3-5} \cmidrule(l){6-8}
            & $\mathcal{A}_{tc}$ (\%, $\uparrow$)      & $76.6$  & $89.2$      & $\bm{100.0}$  & $2.18$  & $83.2$     & $\bm{100.0}$\\
            & $\delta$ ($\downarrow$)                  & $\bm{0.025}$ & $0.059$     & $0.095$ & $\bm{0.025}$ & $0.104$     & $0.090$ \\
            & $\varepsilon$ (\%, $\downarrow$)         & $\bm{2.76}$  & $3.10$      & $3.05$  & $14.76$ & $8.63$      & $\bm{6.27}$ \\
            & $\rho_{\mathrm{out}}$ (\%, $\downarrow$) & $9.22$  & $0.05$      & $\bm{0.00}$  & $13.51$ & $0.22$      & $\bm{0.00}$ \\
            & $\rho_{\mathrm{ovl}}$ (\%, $\downarrow$) & $20.46$ & $0.26$      & $\bm{0.00}$  & $16.40$ & $3.23$      & $\bm{0.00}$ \\
        \bottomrule
    \end{tabular}
    \label{tab:input_grouped_comparison}
\end{table*}

\paragraph{Boundary-free baselines.}
On RPLAN, HypergraphFormer attains the lowest GED ($1.62$ vs.\ $2.59$ for HG and $1.95$ for HD) and is the only method whose strict accuracy exceeds the single-digit range, with $\mathcal{A} = 40.9\%$ compared with $6.0\%$ for HG and $16.3\%$ for HD. The room-multiset accuracy is even more decisive: HF reaches $\mathcal{A}_{tc} = 100.0\%$, expected by construction since our procedural add/remove edit (Algorithm~\ref{alg:addremove}, Appendix~\ref{app:editing_tasks_algorithms}) enforces an exact match between predicted and target room multisets; HG, by contrast, satisfies the multiset constraint only $44.2\%$ of the time, and HD reaches $96.7\%$ with markedly higher GED, meaning that even when its set of rooms is correct it frequently fails to match the required access connectivity. HD's deficit from $100\%$ is itself diagnostic: HD allocates a room polygon per input node by construction, so any drop below $100\%$ on $\mathcal{A}_{tc}$ reflects rooms that are obscured by overlapping or degenerate polygons rather than missing rooms. The contrast sharpens on the out of distribution data: on WMR24, HG's and HD's GED nearly doubles ($2.59 \to 3.80$, $1.95 \to 3.78$) and HD's strict accuracy collapses ($16.3\% \to 2.6\%$), while HypergraphFormer slightly \emph{improves} on both metrics ($\mathrm{GED}\ 1.62 \to 1.70$, $\mathcal{A}\ 40.9\% \to 52.5\%$); HF also retains near-perfect $\mathcal{A}_{tc}$ ($99.9\%$), whereas HD drops to $80.0\%$ and HG to $37.1\%$. The per-bin breakdown in Appendix~\ref{app:per_bin_results} (Table~\ref{tab:per_bin_comparison}) confirms that HF's advantage \emph{widens} with apartment complexity on both datasets, whereas HG's and HD's GED degrade steadily as the room count grows.

\paragraph{Boundary-constrained baselines.}
On RPLAN, HF reaches the multiset target by design ($\mathcal{A}_{tc} = 100.0\%$) while DP and iPLAN stay at $89.2\%$ and $76.6\%$. iPLAN attains the lowest $\delta$ ($0.025$) and lowest $\varepsilon$ ($2.76\%$) but at the cost of severe tiling violations ($\rho_{\mathrm{out}} = 9.22\%$ of total apartment area placed outside the boundary, $\rho_{\mathrm{ovl}} = 20.46\%$ inter-room overlap); DP improves substantially on tiling ($\rho_{\mathrm{out}} = 0.05\%$, $\rho_{\mathrm{ovl}} = 0.26\%$) at a small cost in $\delta$; and HF achieves $\rho_{\mathrm{out}} = \rho_{\mathrm{ovl}} = 0\%$ by design since BSP-derived rooms tile the apartment exactly with no gaps or overlaps, while landing essentially on top of DP on $\varepsilon$ ($3.05\%$ vs.\ $3.10\%$) and only modestly behind iPLAN. Out of distribution, the asymmetry is more pronounced. iPLAN's structural fidelity collapses ($\mathcal{A}_{tc}$ from $76.6\% \to 2.18\%$), DP also degrades but less sharply ($\mathcal{A}_{tc}$ from $89.2\% \to 83.2\%$), while HF maintains $\mathcal{A}_{tc} = 100\%$ and exact tiling ($\rho_{\mathrm{out}} = \rho_{\mathrm{ovl}} = 0\%$). The geometric ranking on $\delta$ and $\varepsilon$ also moves in HF's favor: HF's $\varepsilon$ ($6.27\%$) becomes the lowest of the three, and HF's $\delta$ ($0.090$) overtakes DP ($0.104$). iPLAN's nominal $\delta = 0.025$ remains the lowest in the column but, read alongside its $\mathcal{A}_{tc} = 2.18\%$ and double-digit $\rho_{\mathrm{out}}/\rho_{\mathrm{ovl}}$, reflects per-room shape statistics computed on the small minority of plans whose room set is recovered at all and so is not a like-for-like comparison with HF and DP. The combined picture is that HF wins decisively on the structural and tiling metrics that crucially determine the validity of a floor plan, gives up only a small in-distribution constant on $\delta$ where the BSP tiling constraint slightly limits per-room squareness (a trade-off further analyzed in Appendix~\ref{app:editing_tasks_aggregate}), and overtakes both DP and iPLAN on those same geometric metrics under distribution shift.

\paragraph{Data efficiency.}
\label{sec:data_efficiency}

We re-run supervised fine-tuning of HypergraphFormer (from the same pretrained LLM checkpoint, with all hyperparameters held fixed) on progressively smaller random subsets of the RPLAN training set ($1{,}000$, $5{,}000$, $10{,}000$, and $25{,}000$ samples, against the full set of $\sim\!50{,}000$), and apply the same procedural post-processing pipeline at evaluation. Table~\ref{tab:data_efficiency_summary} reports GED and GED accuracy $\mathcal{A}$ on RPLAN and on the out-of-distribution WMR24 test set, alongside HouseGAN++ and HouseDiffusion, both of which are trained on the full RPLAN dataset.

\begin{wraptable}{r}{0.38\linewidth}
    \vspace{-0.0em}
    \centering
    \caption{Training data efficiency of HypergraphFormer. GED ($\downarrow$) and GED accuracy $\mathcal{A}$ ($\uparrow$, \%) for RPLAN vs.\ WMR24 compared with HouseGAN++ and HouseDiffusion trained on full dataset.}
    \label{tab:data_efficiency_summary}
    \footnotesize
    \setlength{\tabcolsep}{4pt}
    \begin{tabular}{@{}l cc cc@{}}
        \toprule
        & \multicolumn{2}{c}{\textbf{RPLAN}} & \multicolumn{2}{c@{}}{\textbf{WMR24}} \\
        \cmidrule(lr){2-3} \cmidrule(l){4-5}
        Training & GED & $\mathcal{A}$ & GED & $\mathcal{A}$ \\
        \midrule
        Ours-Full Dataset & $1.62$ & $40.9$ & $1.70$ & $52.5$ \\
        Ours--25{,}000      & $1.97$ & $33.1$ & $2.11$ & $43.3$ \\
        Ours--10{,}000      & $2.97$ & $17.4$ & $2.79$ & $27.1$ \\
        Ours--5{,}000       & $3.68$ & $9.9$  & $3.24$ & $21.9$ \\
        Ours--1{,}000       & $4.69$ & $4.3$  & $4.03$ & $12.3$ \\
        \midrule
        HouseGAN++ & $2.59$ & $6.0$  & $3.80$ & $8.5$ \\
        HouseDiffusion & $1.95$ & $16.3$ & $3.78$ & $2.6$ \\
        \bottomrule
    \end{tabular}
\end{wraptable}

On RPLAN, HypergraphFormer trained on only $25{,}000$ samples (half the original training data) already matches HouseDiffusion's GED ($1.97$ vs.\ $1.95$) while doubling its strict accuracy ($33.1\%$ vs.\ $16.3\%$), and substantially exceeds HouseGAN++ on both metrics. The data-efficiency picture is even sharper out of distribution: on WMR24, HypergraphFormer trained on just $1{,}000$ samples, roughly $1.8\%$ of the data used by the baselines, already attains $\mathcal{A} = 12.3\%$, exceeding both HouseGAN++ ($8.5\%$) and HouseDiffusion ($2.6\%$), and at $5{,}000$ samples it more than doubles HouseGAN++ ($21.9\%$ vs.\ $8.5\%$) and roughly tenfolds HouseDiffusion ($21.9\%$ vs.\ $2.6\%$) while also achieving a lower GED than either. These results indicate that the hypergraph representation, combined with LLM priors, lets HypergraphFormer reach state-of-the-art performance from a small fraction of the training data the baselines require, with the gains amplified rather than diminished under distribution shift. Per room-count breakdowns are reported in Appendix~\ref{app:per_bin_results}.

\paragraph{Procedural editing pipeline.}
\label{sec:ablation_studies}
We ablate the post-generation procedural-editing pipeline (Add/Remove Rooms $\to$ Pick Orientation $\to$ Optimize $\delta$ and $\varepsilon$) on both RPLAN and WMR24. Three observations stand out. \emph{Add/Remove Rooms} raises $\mathcal{A}_{tc}$ from $\approx\!75\%$ to $\geq\!99.8\%$ by construction (Algorithm~\ref{alg:addremove}) and pulls aggregate GED from $1.99$ to $1.72$ on RPLAN and from $1.97$ to $1.72$ on WMR24; \emph{Pick Orientation} then reduces $\delta$ by roughly $25\%$; and the joint $\delta,\varepsilon$ optimizer recovers the lowest $\delta$ ($0.0986$ on RPLAN, $0.070$ on WMR24) without disturbing $\mathcal{A}$. Full per-stage aggregates and detailed discussion are reported in Appendix~\ref{app:editing_tasks_aggregate} (Table~\ref{tab:ablation_studies_summary}); per room-count breakdowns are in Appendix~\ref{app:editing_tasks_per_room}.

\paragraph{Editing via tool calls.}
\label{sec:complex_editing_tasks}

\begin{figure}[!htbp]
    \centering
    \footnotesize
    \vspace{0.85em}
    \setlength{\tabcolsep}{16pt}
    \renewcommand{\arraystretch}{1.0}
    \begin{tabular}{@{}ccc@{}}
        \begin{minipage}[t]{\dimexpr(\columnwidth-4\tabcolsep)/3\relax}\centering
            (a) delete balcony\\[0.45em]
            \includegraphics[width=\linewidth,keepaspectratio]{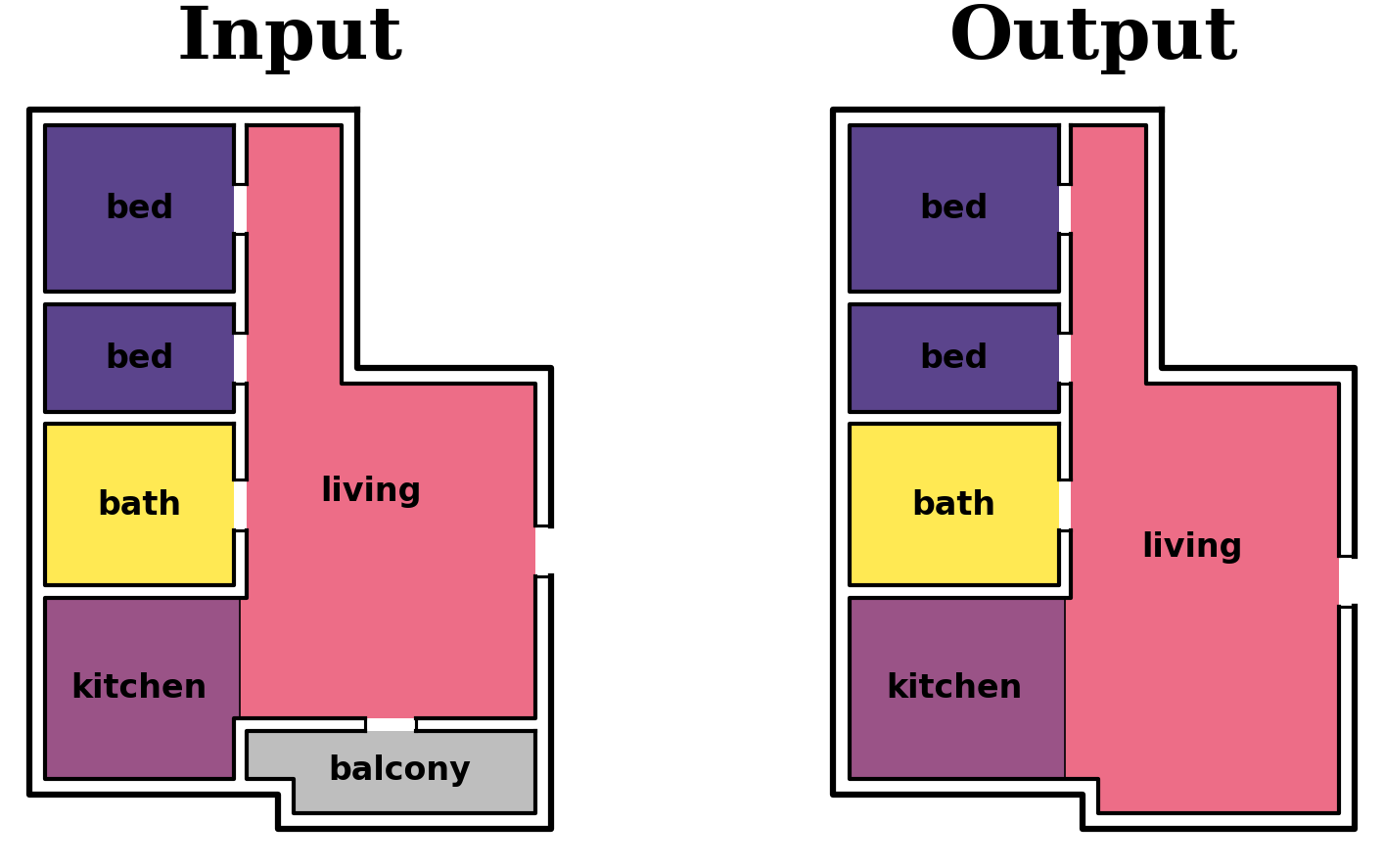}\vspace{2em}
        \end{minipage}
            & \begin{minipage}[t]{\dimexpr(\columnwidth-4\tabcolsep)/3\relax}\centering
            (b) add bathroom by bedroom\\[0.45em]
            \includegraphics[width=\linewidth,keepaspectratio]{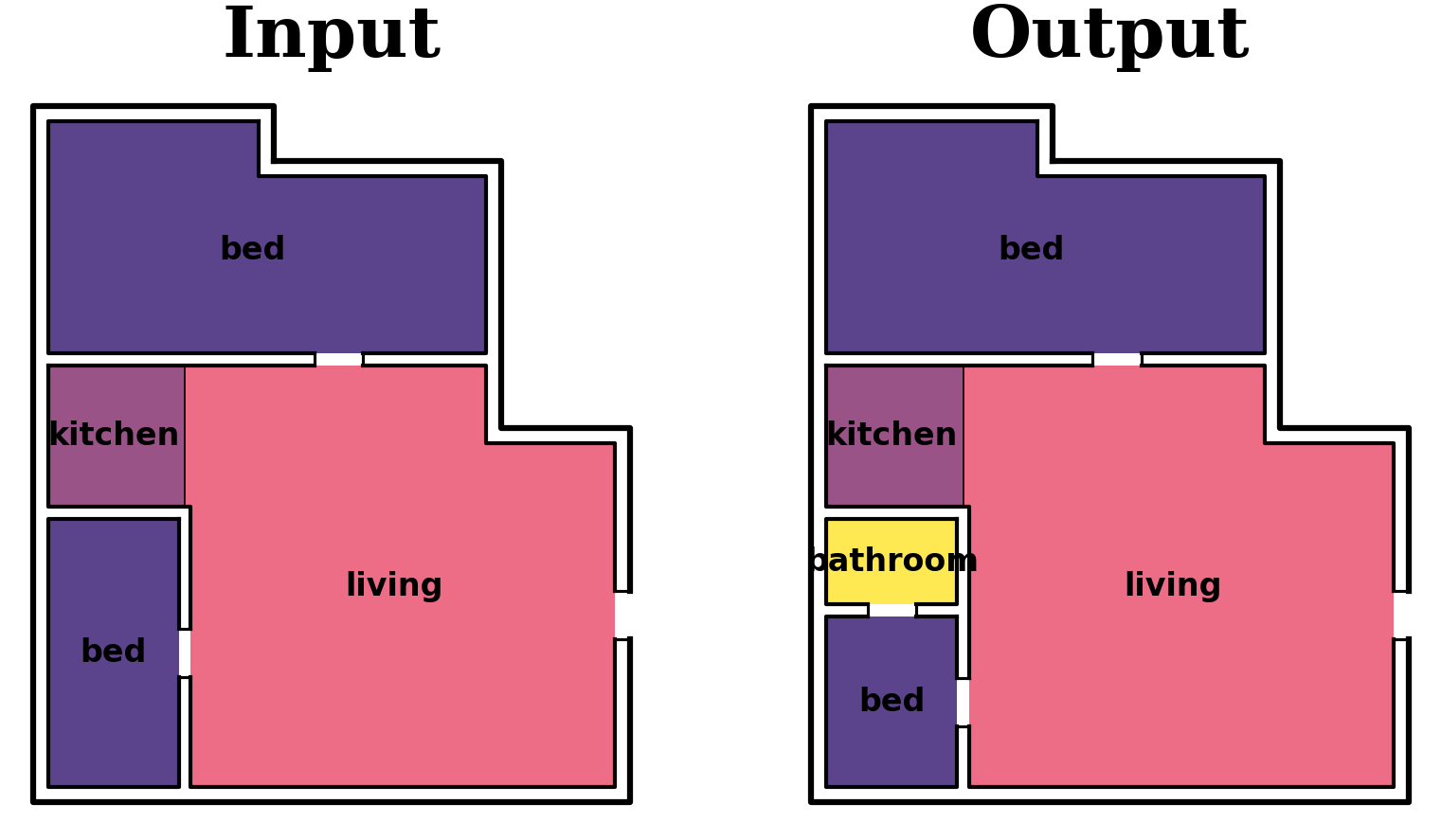}\vspace{2em}
        \end{minipage}
            & \begin{minipage}[t]{\dimexpr(\columnwidth-4\tabcolsep)/3\relax}\centering
            (c) reduce storage\\[0.45em]
            \includegraphics[width=\linewidth,keepaspectratio]{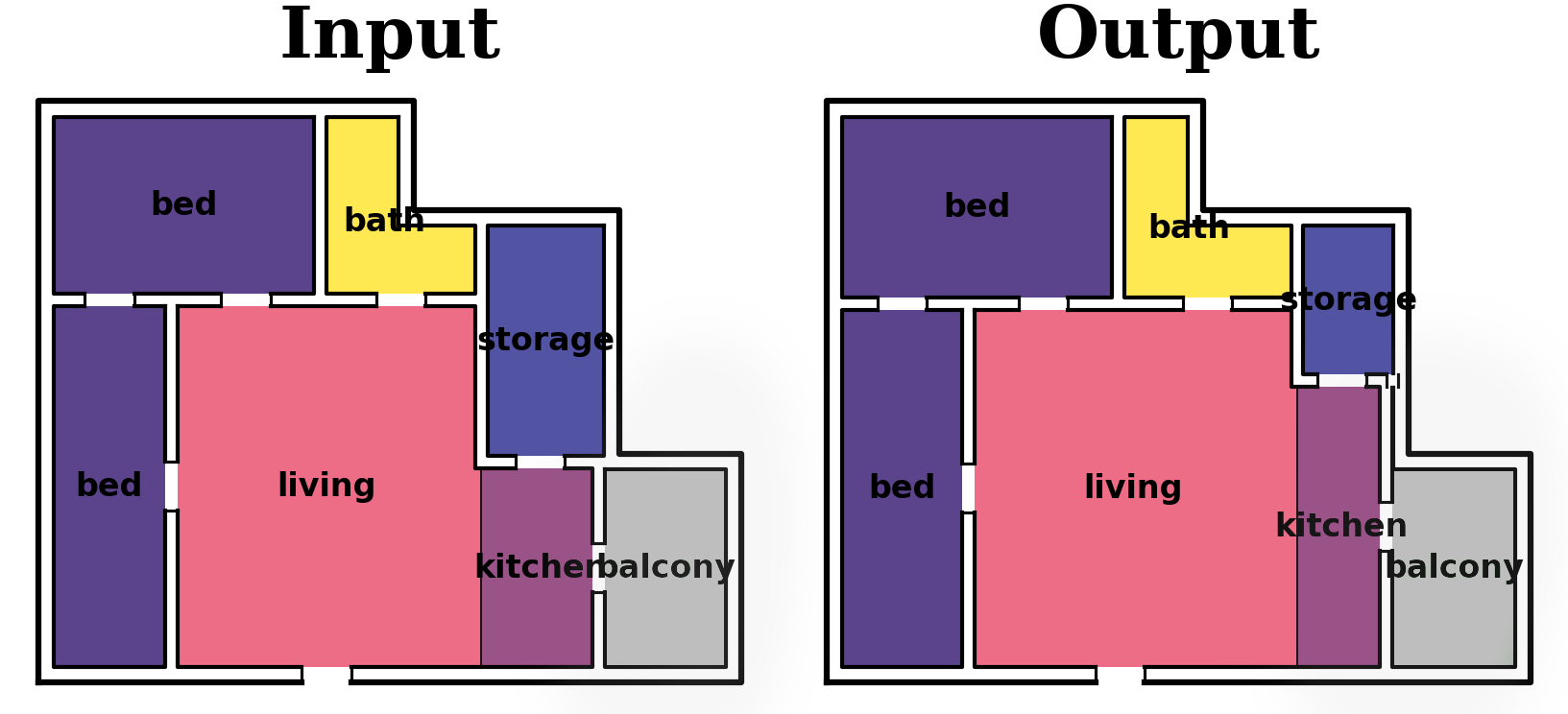}\vspace{2em}
        \end{minipage} \\[4em]
        \begin{minipage}[t]{\dimexpr(\columnwidth-4\tabcolsep)/3\relax}\centering
            (d) freeze part of floor plan\\[0.45em]
            \includegraphics[width=\linewidth,keepaspectratio]{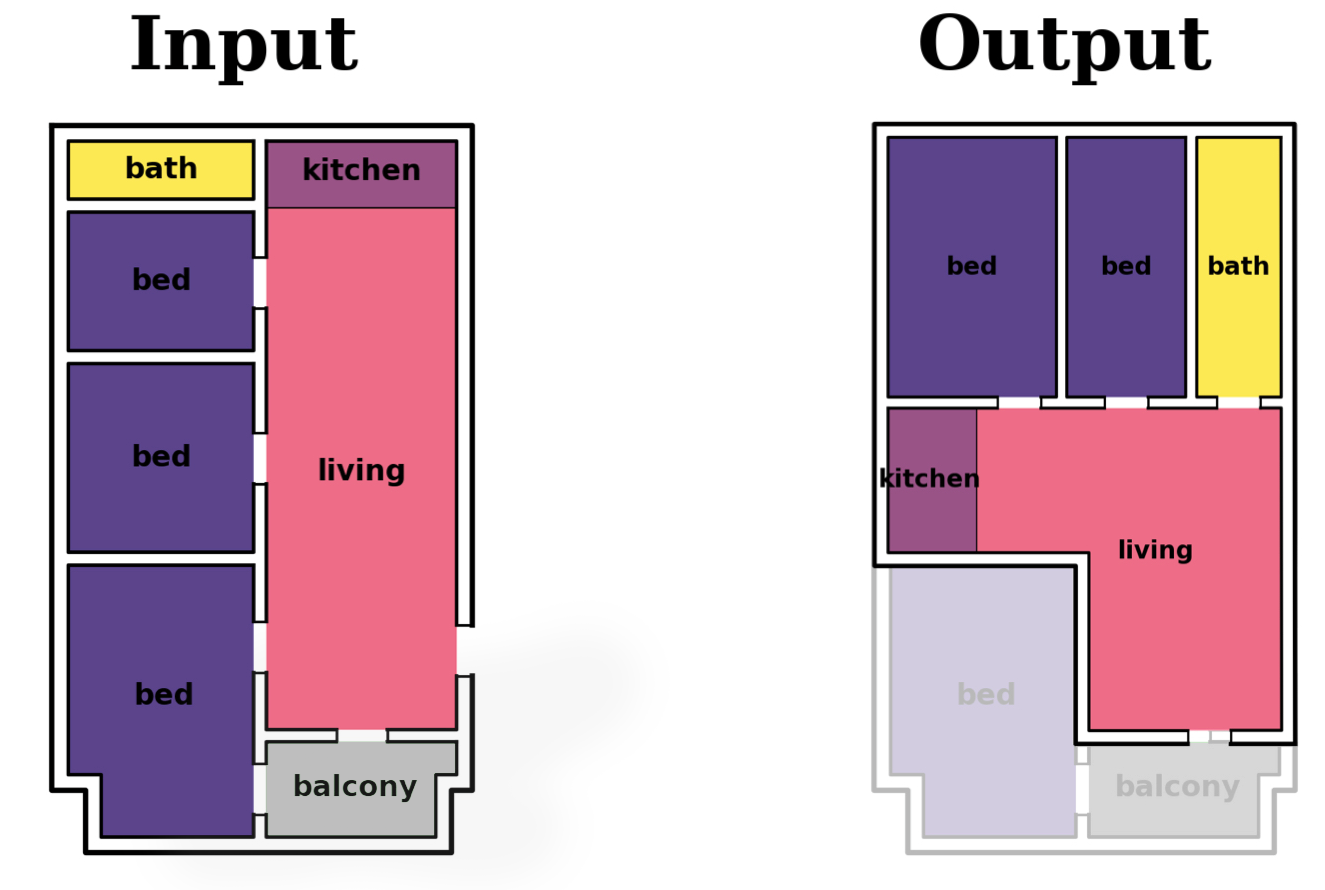}\vspace{2em}
        \end{minipage}
            & \begin{minipage}[t]{\dimexpr(\columnwidth-4\tabcolsep)/3\relax}\centering
            (e) swap kitchen and bedroom\\[0.45em]
            \includegraphics[width=\linewidth,keepaspectratio]{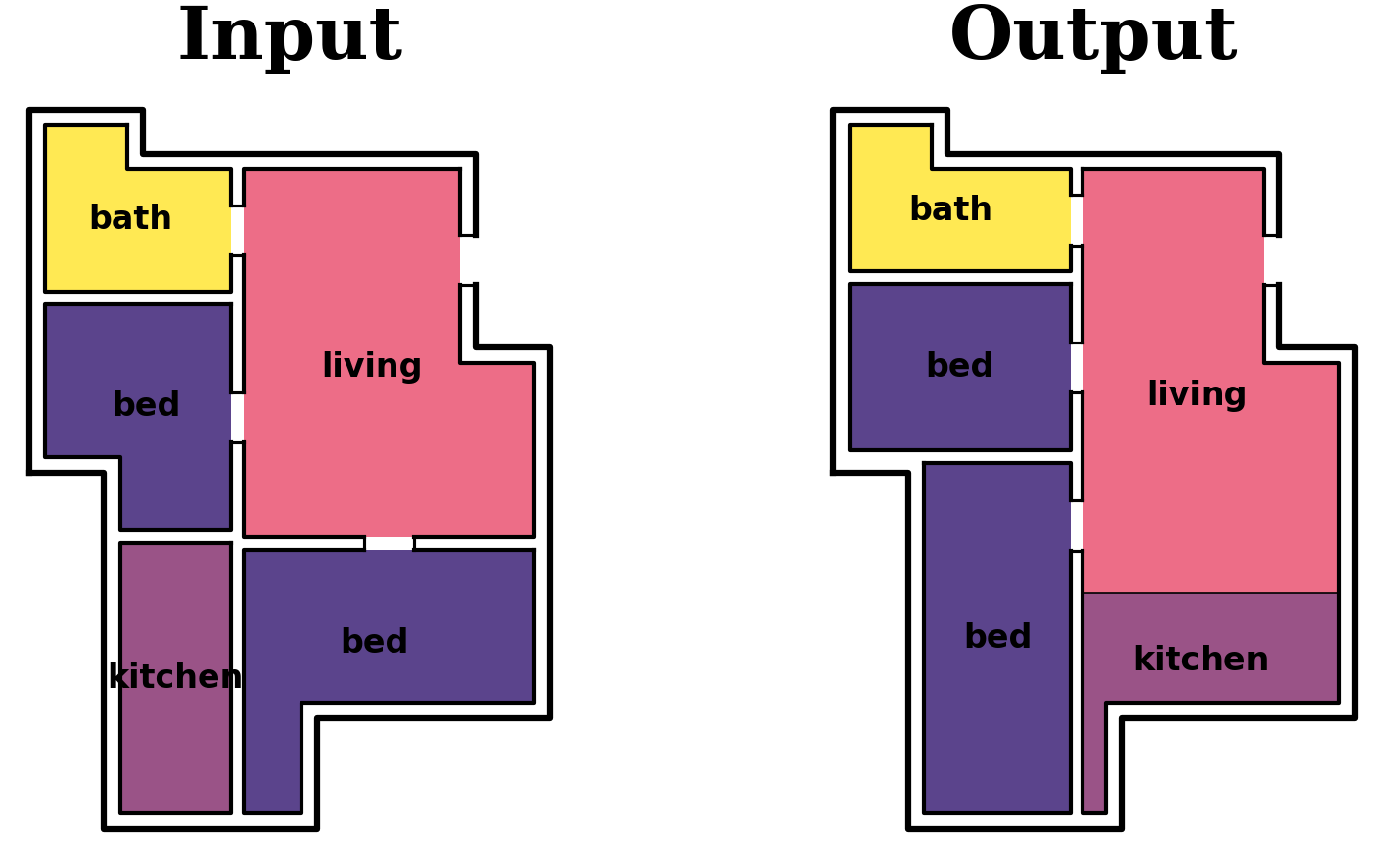}\vspace{2em}
        \end{minipage}
            & \begin{minipage}[t]{\dimexpr(\columnwidth-4\tabcolsep)/3\relax}\centering
            (f) rotate 90\\[0.45em]
            \includegraphics[width=\linewidth,keepaspectratio]{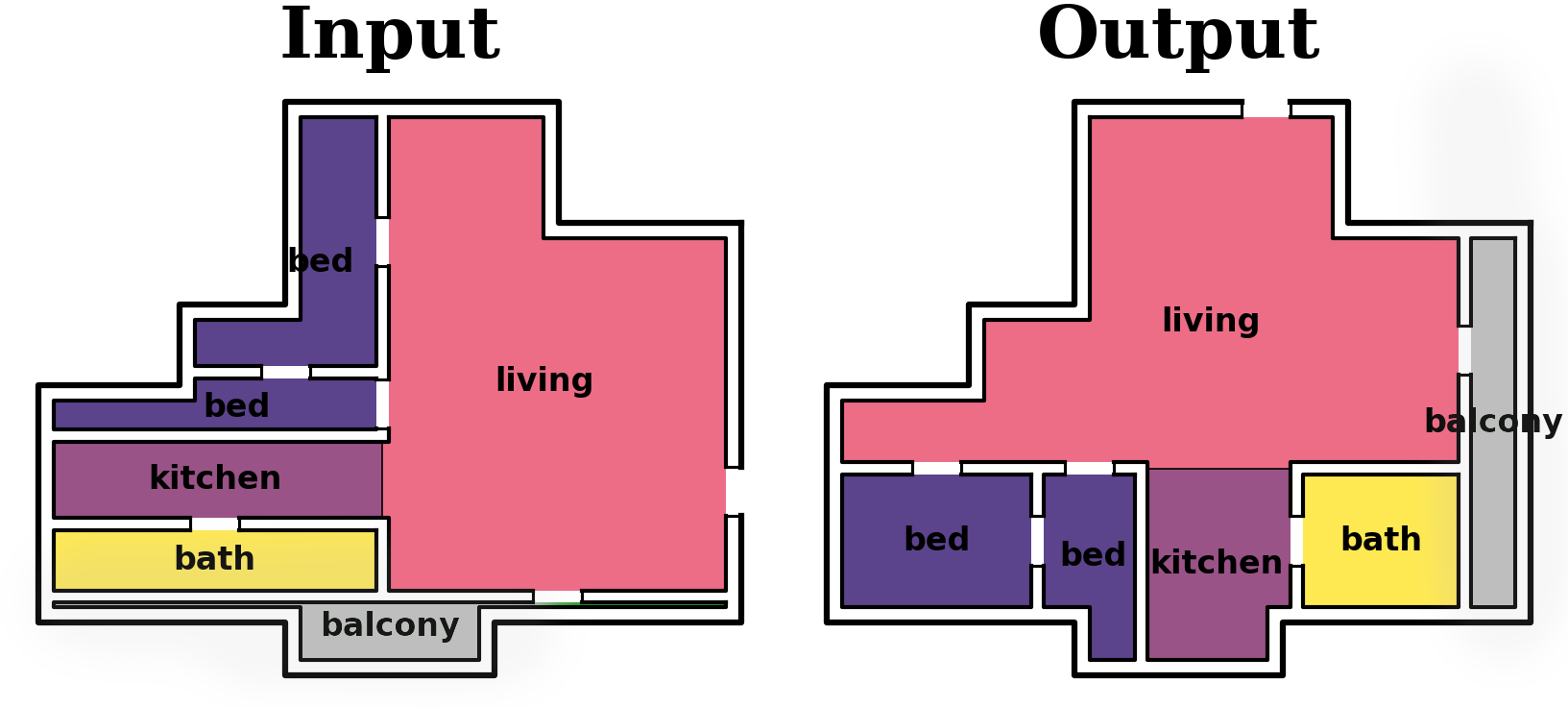}\vspace{2em}
        \end{minipage} \\[4em]
        \begin{minipage}[t]{\dimexpr(\columnwidth-4\tabcolsep)/3\relax}\centering
            (g) flip along x-axis\\[0.45em]
            \includegraphics[width=\linewidth,keepaspectratio]{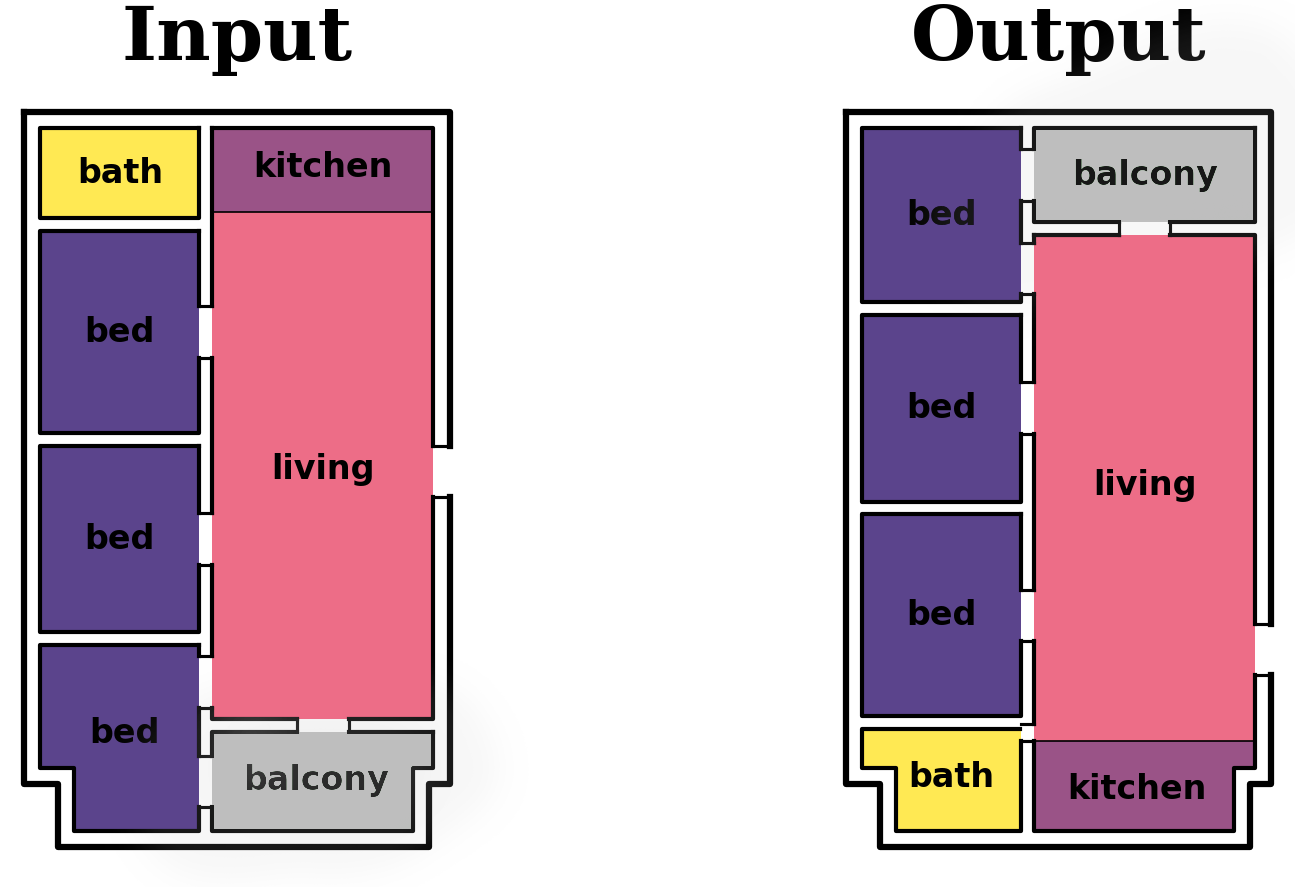}\vspace{2em}
        \end{minipage}
            & \begin{minipage}[t]{\dimexpr(\columnwidth-4\tabcolsep)/3\relax}\centering
            (h) move entrance to the right\\[0.45em]
            \includegraphics[width=\linewidth,keepaspectratio]{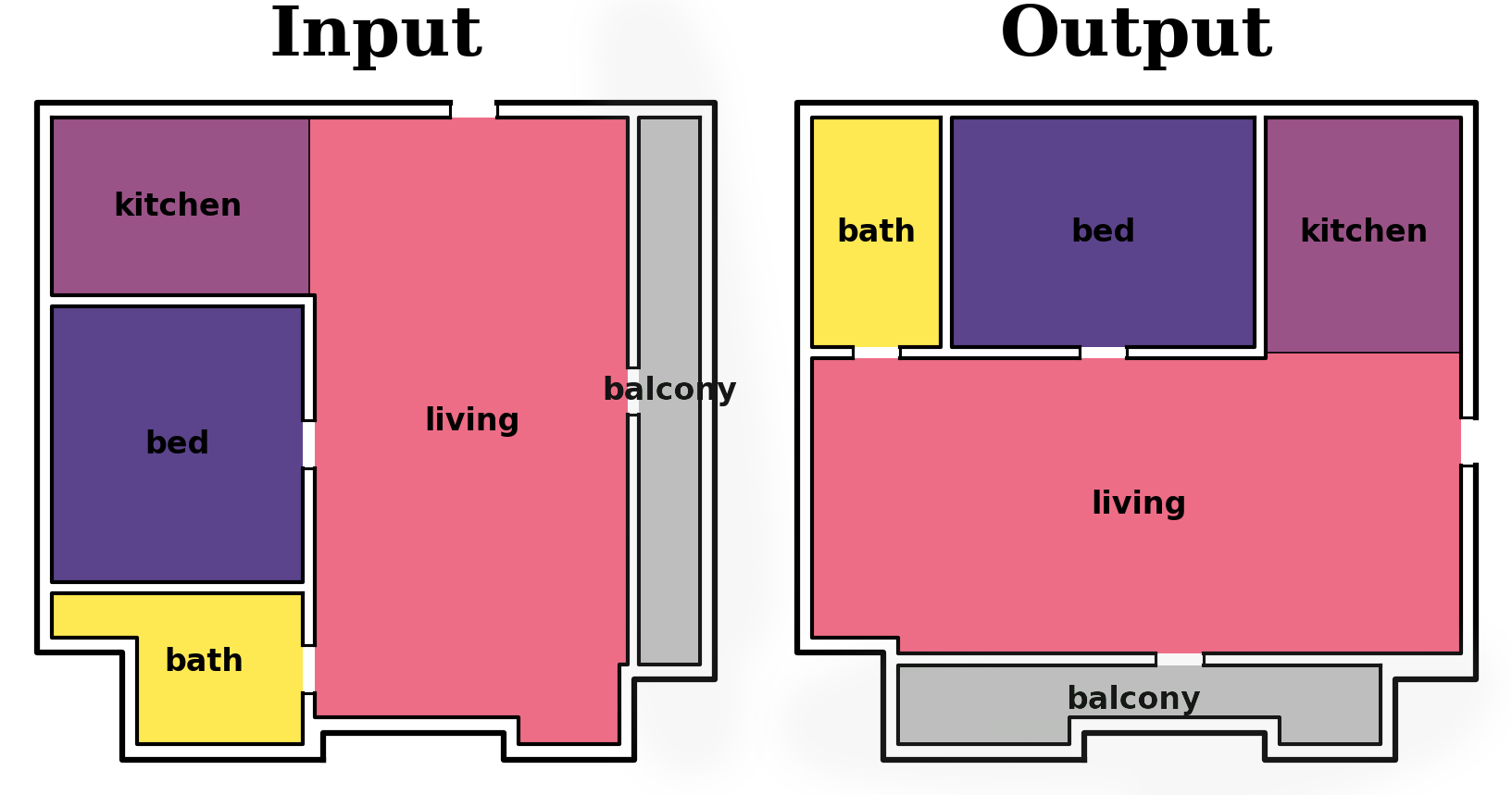}\vspace{2em}
        \end{minipage}
            & \begin{minipage}[t]{\dimexpr(\columnwidth-4\tabcolsep)/3\relax}\centering
            (i) orient bedroom to top\\[0.45em]
            \includegraphics[width=\linewidth,keepaspectratio]{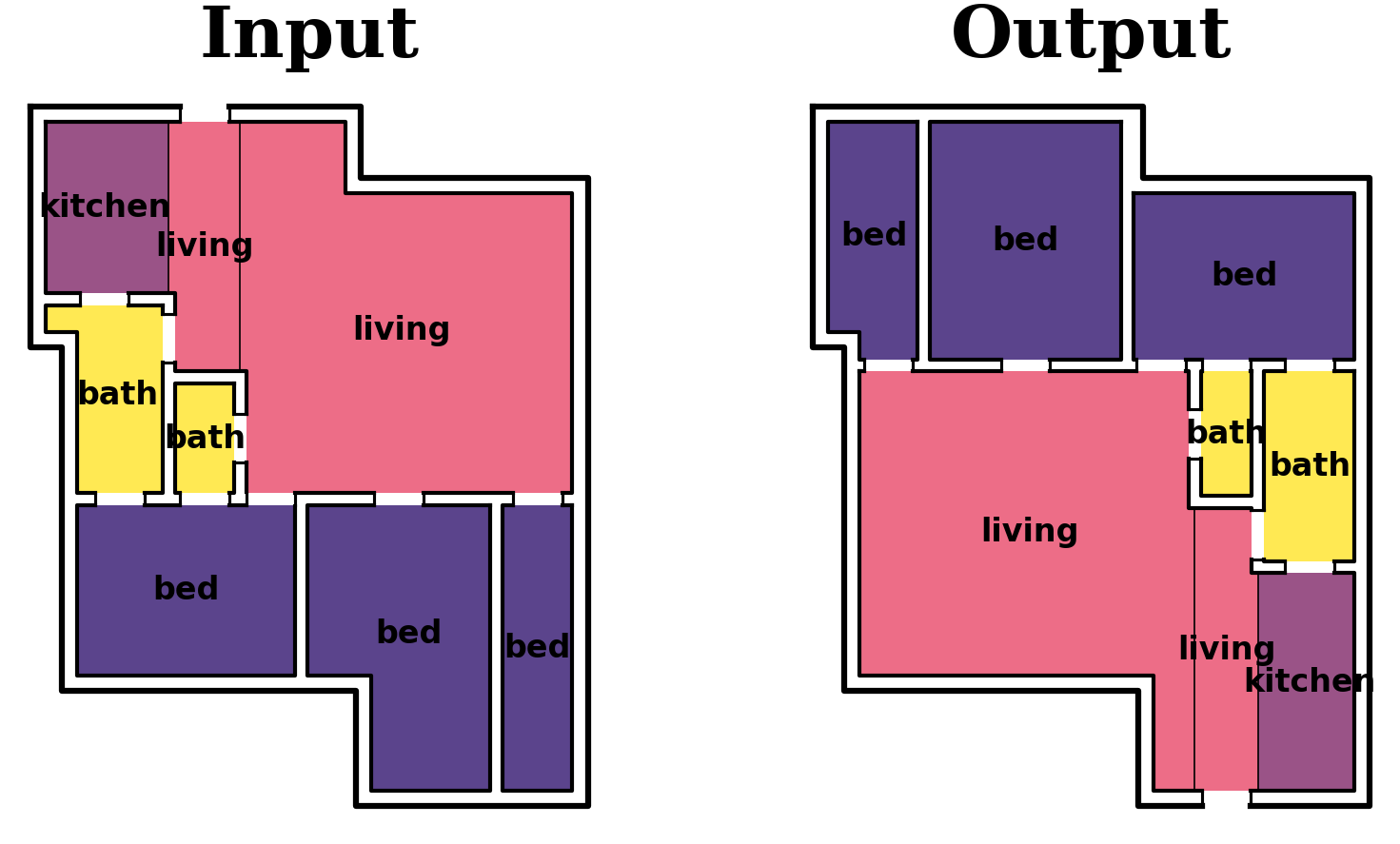}\vspace{2em}
        \end{minipage} \\[4em]
    \end{tabular}
    \caption{Illustrative hypergraph edits. The title of each panel names the corresponding edit command.}
    \label{fig:complex_editing_tasks_examples}
\end{figure}

\begin{figure}[!htbp]
    \centering
    \footnotesize
    \vspace{0.3em}
    \setlength{\tabcolsep}{0pt}
    \renewcommand{\arraystretch}{1.0}
    \begin{tabular}{@{}cccccc@{}}
        \begin{minipage}[c]{\dimexpr(\columnwidth-7\tabcolsep)/6\relax}\centering
            \includegraphics[width=\linewidth,keepaspectratio]{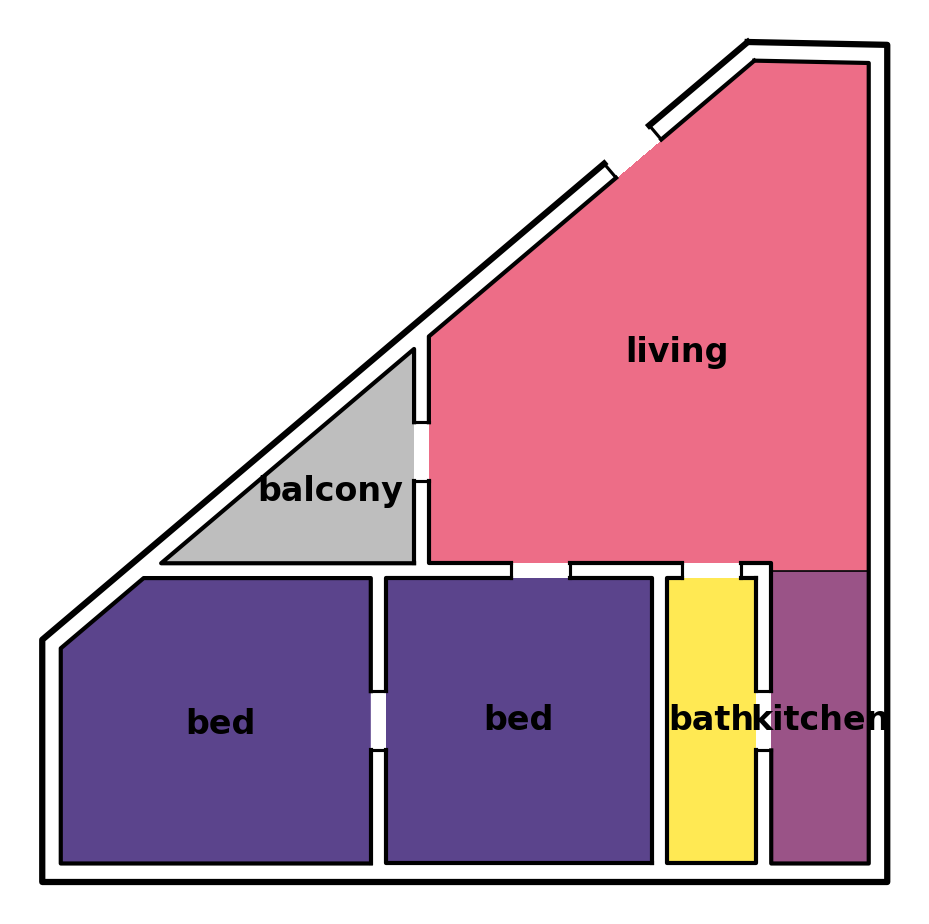}
        \end{minipage}
            & \begin{minipage}[c]{\dimexpr(\columnwidth-7\tabcolsep)/6\relax}\centering
            \includegraphics[width=\linewidth,keepaspectratio]{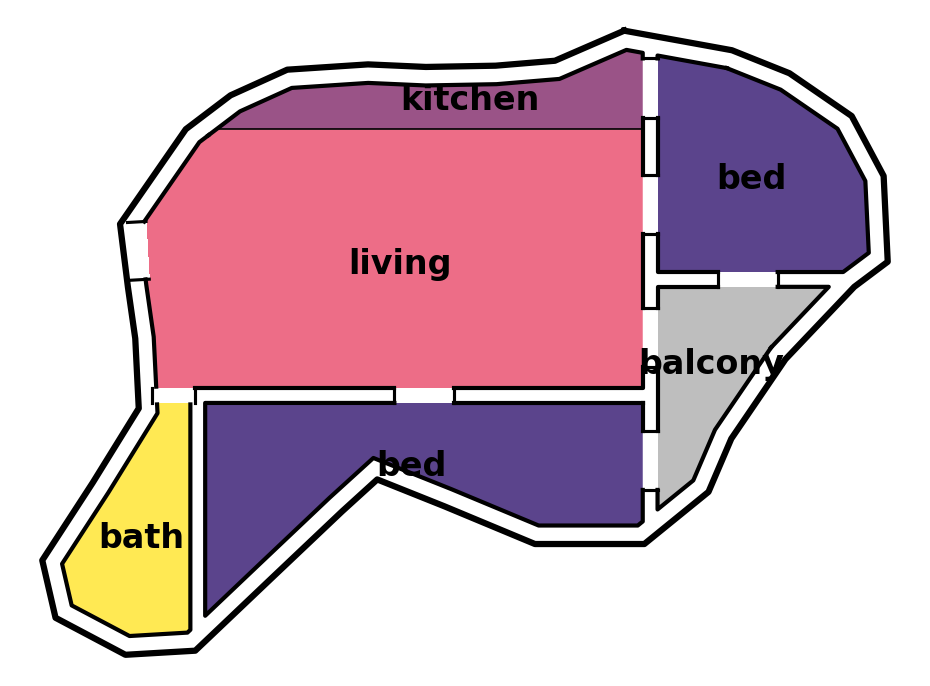}
        \end{minipage}
            & \begin{minipage}[c]{\dimexpr(\columnwidth-7\tabcolsep)/6\relax}\centering
            \includegraphics[width=\linewidth,keepaspectratio]{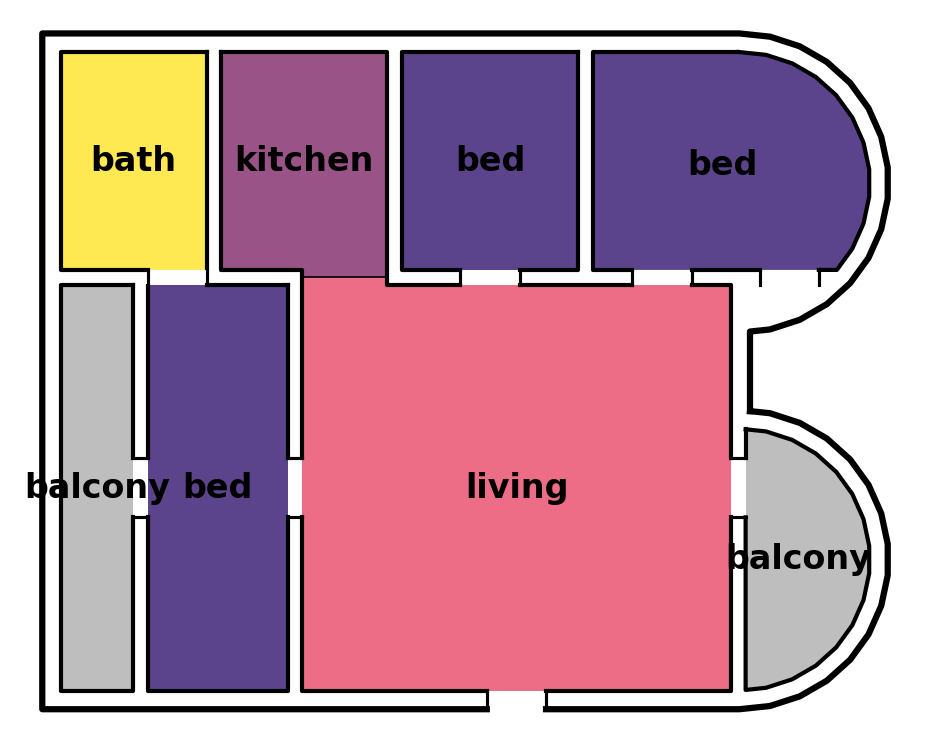}
        \end{minipage}
            & \begin{minipage}[c]{\dimexpr(\columnwidth-7\tabcolsep)/6\relax}\centering
            \includegraphics[width=\linewidth,keepaspectratio]{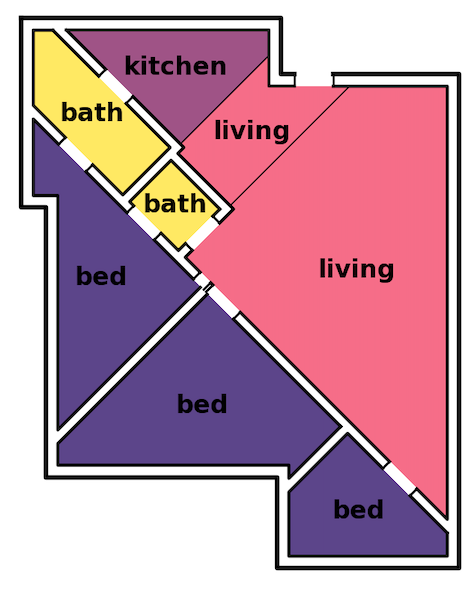}
        \end{minipage}
            & \begin{minipage}[c]{\dimexpr(\columnwidth-7\tabcolsep)/6\relax}\centering
            \includegraphics[width=\linewidth,keepaspectratio]{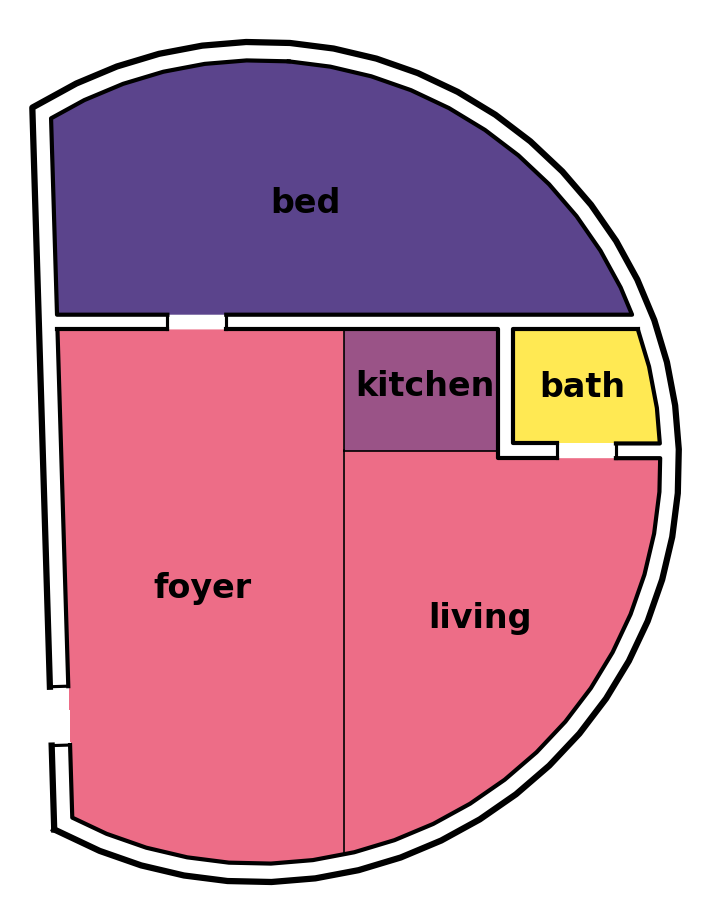}
        \end{minipage}
            & \begin{minipage}[c]{\dimexpr(\columnwidth-7\tabcolsep)/6\relax}\centering
            \includegraphics[width=\linewidth,keepaspectratio]{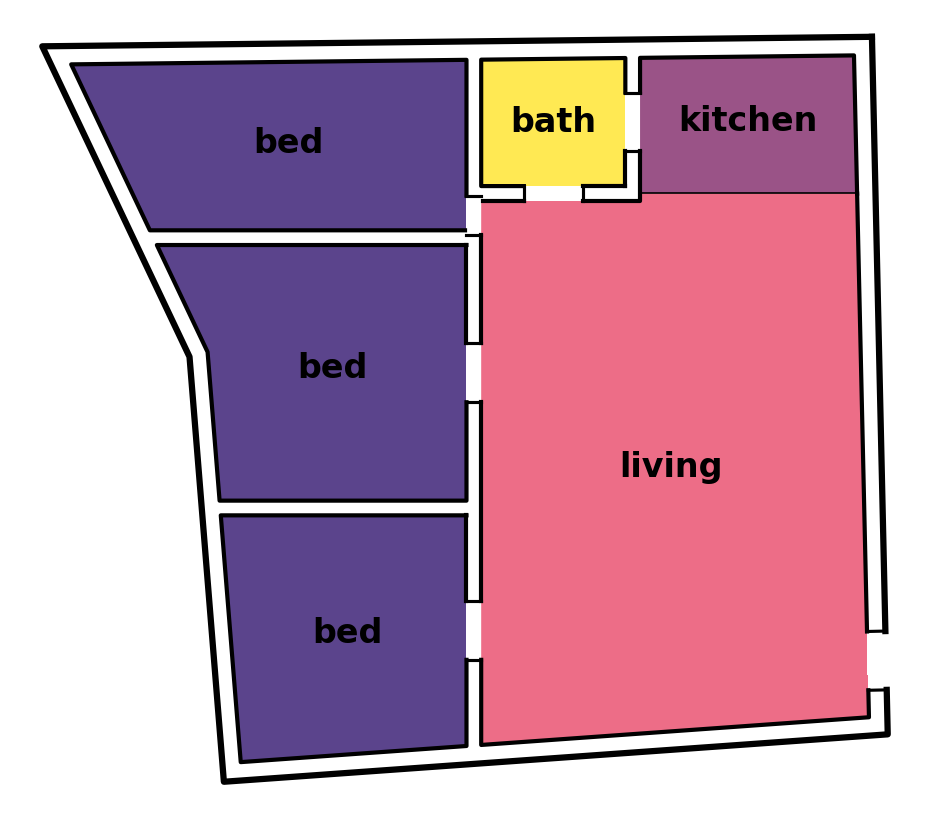}
        \end{minipage} \\
    \end{tabular}
    \caption{Examples of HypergraphFormer outputs fitted to non-Manhattan apartment boundaries.}
    \label{fig:nonmanhattan_layouts}
\end{figure}

Beyond the three procedural stages used in the post-processing pipeline, the hypergraph format admits a broader family of edits that an LLM can invoke as tool calls in response to a designer's instructions, either to improve the quality of a generated layout or to repurpose it for downstream design exploration. Fig.~\ref{fig:complex_editing_tasks_examples} illustrates representative interior edits (deleting, adding, swapping, resizing, reorienting, and freezing rooms or sub-regions), and Fig.~\ref{fig:nonmanhattan_layouts} shows the complementary case where the user redraws the apartment boundary to an arbitrary, non-Manhattan shape and a generated hypergraph is re-fitted to it. Each edit in Fig.~\ref{fig:complex_editing_tasks_examples} acts directly on the BSP tree and access graph and is deterministic, so it composes cleanly with the procedural pipeline above. Per-edit definitions and sample tool calls are provided in Appendix~\ref{app:complex_editing_tasks}.

These operations are intentionally simple and deterministic, yet they already enable substantial structural and geometric changes to a generated layout. The hypergraph representation suggests further opportunities for higher-level editing, such as learning a policy that selects and composes edits in response to a designer's natural-language brief or to downstream performance objectives.

\FloatBarrier

\paragraph{Discussion.}
Our experiments suggest that what unlocks editability, data efficiency, and out-of-distribution generalization for floor plan generation is not model scale but \emph{representation}. By emitting a hypergraph rather than pixels or per-room polygons, the LLM is asked to predict a small, structured object whose constraints (BSP topology, leaf-only connectivity, area conservation) it can match exactly, instead of a high-dimensional geometry whose validity it can only approximate. The same compactness that makes the format easy to learn also makes it easy to edit: a few-token change to the BSP tree or access graph corresponds to a meaningful structural edit, with no re-inference pass. The trade-off is geometric expressivity --- the BSP tiling constraint gives up modest per-room squareness on $\delta$ in exchange for exact boundary tiling, exact room-multiset accuracy, and a representation that an LLM can manipulate compositionally.

We further elaborate on the limitations of this hypergraph formulation, together with the directions for future work that they motivate, in Appendix~\ref{app:future_work}.
\FloatBarrier


\section{Conclusion}
\label{sec:conclusion}
We presented \emph{HypergraphFormer}, a representation-centric framework for editable floor plan generation that decouples interior layout structure from apartment geometry through a hypergraph formulation learned with large language models. The framework rests on four pillars that together distinguish it from prior work: (i) \emph{out-of-distribution generalization}, with an RPLAN-trained model surpassing rasterized and vectorized baselines on architect-designed plans whose boundary shapes and conventions differ markedly from the training distribution; (ii) \emph{data efficiency}, reaching baseline accuracy from a small fraction of the training data they require, with the largest reductions observed under distribution shift; (iii) a \emph{procedural-editing pipeline} of deterministic, LLM-free post-processing operations (room add/remove, orientation selection, and gradient-descent refinement of area splits) that compose with the generated hypergraph; and (iv) \emph{tool-call edits}, in which an LLM invokes named operations directly on the BSP tree and access graph to apply higher-level structural and geometric changes. To support evaluation in this setting and to enable reproduction and extension of our work, we release \emph{WMR24}, a curated benchmark of architect-designed floor plans, alongside our full training, inference, dataset-conversion, and procedural-editing code. Together, these results show that explicitly learning a structured, boundary-independent hypergraph, rather than relying on model complexity or image synthesis, is what unlocks scalability, editability, and robustness, and they point to representation as a central lever for integrating large language models with human-interpretable design abstractions.

\bibliography{main}
\bibliographystyle{unsrtnat}


\appendix
\section{Related work}
\label{sec:related}
Automatic floor plan generation has been an active area of research even before the advent of deep learning. \citet{bayesian} formulate the problem as an optimization task and train a Bayesian network for the purpose. With the advent of deep learning, numerous approaches have since been proposed. Graph2Plan~\cite{hu2020graph2plan} combines a graph neural network with a retrieve-and-adjust paradigm to generate floor plans from layout graphs and building boundaries; however, its reliance on retrieving template graphs from the RPLAN dataset makes it less applicable to data-scarce settings of real-world designs. GAN-based approaches have also been developed, in which each room is represented as a binary mask and the masks are combined to produce the final layout~\cite{housegan}. This line of work has been extended through iterative refinement~\cite{housegan++}, by replacing the input spatial \emph{adjacency} graph with functional \emph{access}-graph constraints that more effectively capture user-specified design preferences, and through further developments such as graph-structured masked autoencoders~\cite{MaskPlan}. A major limitation of rasterized representations is that the generated set of masks requires non-trivial post-processing and integration to be converted into a usable floor plan. HouseDiffusion~\cite{housediffusion} formulates floor plan generation in a vectorized representation and uses diffusion models to solve the task; its key novelty is the use of discretized final 2D coordinates to establish incidence relationships. HouseTune~\cite{housetune} proposes a two-stage framework that leverages LLMs to interpret user design specifications expressed in natural language and generate an initial layout, which is subsequently refined by a diffusion model into a realistic floor plan. Although the aforementioned models achieve promising performance in terms of quantitative metrics and visual realism, they typically generate the building boundary implicitly during synthesis, which limits explicit control over the apartment footprint and may be misaligned with user design preferences. iPLAN~\cite{iplan} adopts a raster-based generative model constrained by the input boundary; however, the rasterized nature of the model limits its ability to generate non-Manhattan or complex floor plans. DiffPlanner~\cite{DiffPlanner} represents floor plans purely in vector space, with rooms encoded as top-left and bottom-right coordinate pairs inside given bounds, yielding simple but boundary-constrained plans that still require post-processing. Adjacency graphs have also been widely used as a way to analyse and compare floor plans~\cite{engelenburgGKN_2025}, with rooms as nodes connected by edges that denote, e.g., a shared door. Although such adjacency graphs have been combined with Graph Neural Networks (GNNs) for floor plan tasks, they cannot directly generate rooms and instead require secondary geometric procedures~\cite{Lu2025DeepEdgeAwareGNN, promptsToLayouts}.

Table~\ref{tab:related_work_comparison} compares representative deep-learning floor plan generators across input/output representations, editability, and extra capabilities. We organize the discussion around two themes: \emph{representation} and \emph{editability and generalization}.

\paragraph{Input and output representations.} Most baselines fall into two camps. The first camp consumes an \emph{adjacency graph} (rooms as nodes, shared-wall edges) together with a building boundary and produces either a raster mask~\cite{housegan,Tang2023GraphTransformer,MaskPlan,hu2020graph2plan} or a vector floor plan~\cite{wallplan,DiffPlanner,GSDiff}; the second camp consumes an \emph{access graph} (rooms as nodes, door-connection edges)~\cite{housegan++,housediffusion} or a natural-language description~\cite{housetune} and produces a vector or raster output. The distinction between adjacency and access graph is consequential: an adjacency graph already encodes the topological skeleton of the layout (every wall-sharing pair), so methods conditioned on it effectively receive a near-complete description of the partitioning before generation begins. An access graph is strictly less informative (door connectivity is a sparse subset of wall adjacency) and forces the model to infer the geometric scaffold from the door connectivity alone. HypergraphFormer is the only method whose primary input is an access graph and whose output is itself a graph-structured representation rather than a raster or per-room polygon list. Because the textual access graph can be degraded by removing edge information, HypergraphFormer additionally supports trivial coarser conditioning, such as room-count-only or room-type-only prompts, without architectural changes; methods conditioned on a fully specified adjacency cannot easily fall back to such sparser specifications.

\paragraph{Editability and generalization.} The second half of the table makes the cost of editability explicit. Among the baselines, only methods that have built dedicated training-time machinery offer non-trivial fine-grained control: MaskPLAN's dynamic masked autoencoder over multi-modal layout attributes~\cite{MaskPlan}, iPLAN's reverse-engineered stage-by-stage Markov chain~\cite{iplan}, DiffPlanner's three-stage cascaded diffusion with partial-condition dropout~\cite{DiffPlanner}, and Graph2Plan's retrieve-and-adjust pipeline~\cite{hu2020graph2plan} all introduce non-trivial extra training complexity, and at edit time still require a full re-inference pass through a diffusion process or retrieval+refinement loop. Boundary editability shows the same pattern: every baseline that accepts a user-specified outer boundary does so as conditioning input that requires re-running the full forward pipeline whenever the boundary changes, hence the \pmark{} marks in Table~\ref{tab:related_work_comparison}. Similarly, the three baselines that produce non-axis-aligned geometry achieve this either by augmenting the training set with hand-crafted non-Manhattan examples (HouseDiffusion's Non-Manhattan-RPLAN~\cite{housediffusion}; GSDiff's tilted-balcony-walls augmentation~\cite{GSDiff}) or by zero-shot generalization to non-axis-aligned boundaries at inference time with stated reliability limits (iPLAN's Figure 6~\cite{iplan}); none has a representation that natively expresses arbitrary wall geometry. Finally, every baseline relies on RPLAN ($\sim$80K) or LIFULL (\textgreater100K) for training and reports no out-of-distribution evaluation. In contrast, because HypergraphFormer represents a floor plan as a textual hypergraph that an LLM can directly emit and locally edit, editability is structural rather than auxiliary: adding, removing, or resizing rooms, swapping types, or re-shaping the boundary corresponds to lightweight, online edits to the textual hypergraph rather than another full diffusion or retrieval pass; arbitrary non-axis-aligned boundaries are expressible by construction; and the hypergraph's compositional structure, combined with LLM priors, yields the data efficiency and out-of-distribution generalization we demonstrate empirically.

\paragraph{LLM finetuning.} Modern LLMs are predominantly based on Transformer architectures trained with next-token prediction \cite{vaswani2017attention,radford2018improving,devlin2019bert}. Empirically observed scaling laws suggest that increasing model size, data, and compute tends to yield predictable gains in downstream performance, motivating the development of large-scale foundation models \cite{scalinglaw,hoffmann2022training,shanahan2024talking, openai_chatgpt_2026, google_gemini_2026}. Beyond pretraining, instruction tuning improves editability and adherence to task requirements, while instruction-tuned models (e.g., FLAN) demonstrate strong zero-shot generalization \cite{wei2022finetuned}. In practice, parameter-efficient adaptation methods such as LoRA \cite{hu2021lora} enable effective fine-tuning of pretrained LLMs with limited domain data. This motivates using instruction-fine-tuned LLMs as generators of our hypergraph-based floor plan representation, where prompts explicitly specify the hypergraph constraints. 

\section{Hypergraph representation}
\label{sec:hypergraph}
\begin{figure*}[t]
    \centering
    \includegraphics[width=.9\textwidth]{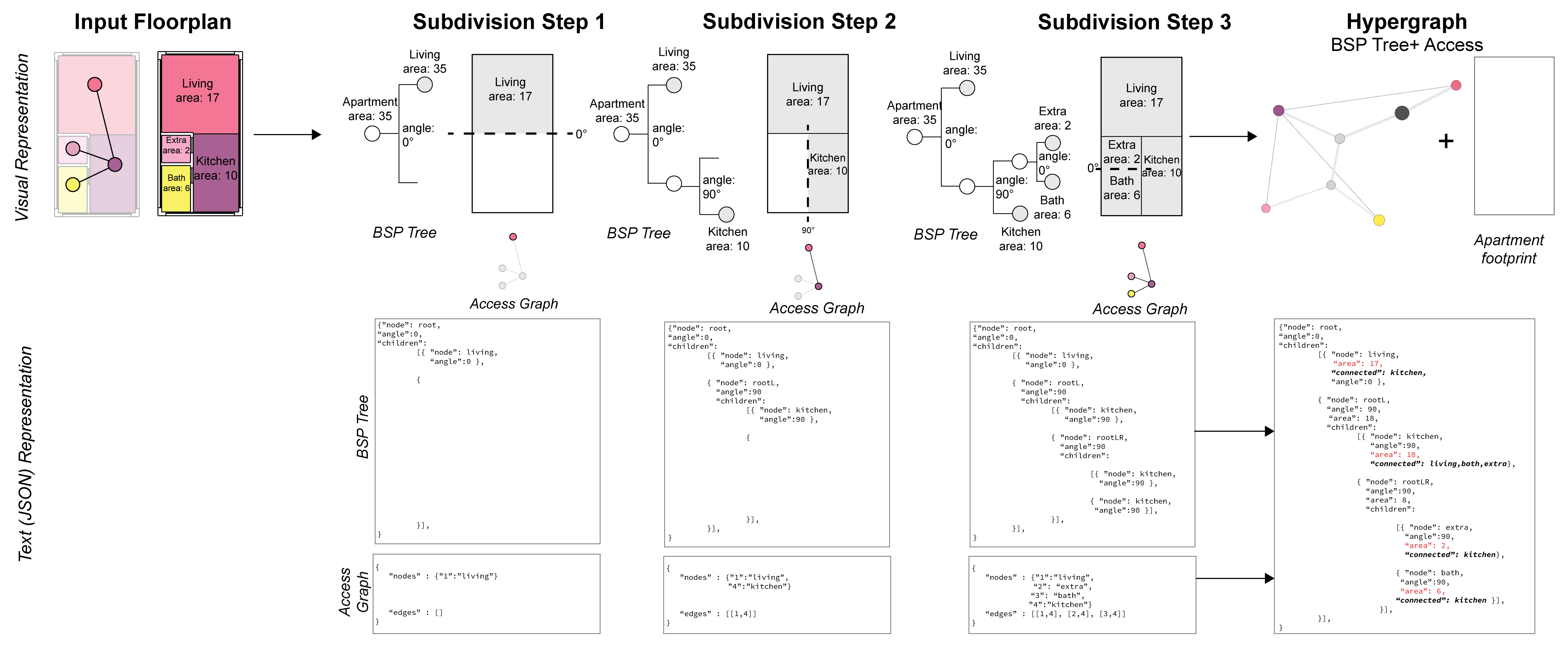}
    \caption{Illustration of the hypergraph representation from both a visual (top) and a data-structure perspective (bottom).}
    \label{fig:hypergraph_representation}
\end{figure*}

As introduced by \citet{hypergraph}, a hypergraph is a reduced-order representation for floor plans. Under this formulation, a floor plan is decomposed into its footprint (or boundary) and a hypergraph (shown in Fig.~\ref{fig:hypergraph_representation}), which may be stored in a structured JSON format. In a hypergraph, each intermediate node denotes a space and encodes semantic attributes such as area and splitting angle, which determine how the space is partitioned. The leaf nodes correspond to individual rooms and store their functions (i.e., room type) together with their connections to other leaf nodes, which together form an access graph. Fig.~\ref{fig:hypergraph_representation} illustrates the process of constructing a hypergraph from a floor plan via binary space partitioning (BSP). In the first step, the root node represents the entire floor plan and is partitioned into two sub-regions, with the splitting angle and area stored at the root. In the second step, one leaf node corresponding to the living area is formed and is no longer subdivided, while the remaining region, represented by an intermediate node, is further partitioned until all remaining rooms are obtained. Finally, an access graph is incorporated into the representation based on doors and open connections between rooms. In the constructed hypergraph, solid edges correspond to BSP tree connections, whereas dashed edges represent the access graph. A key consequence of this reduced representation is that an apartment is decomposed into a boundary-independent hypergraph and an explicit apartment boundary, thereby decoupling the complexity of representing the apartment outline from that of interior layout subdivisions. Known limitations of this representation, together with the directions for future work that they motivate, are discussed in Appendix~\ref{app:future_work}.

\section{Implementation Details}
\label{app:implementation_details}
For supervised fine-tuning we use the Qwen-3-4B-Instruct-2507 checkpoint from Hugging Face~\cite{huggingface}, fine-tuned on 8 NVIDIA A100 GPUs with the Hugging Face Accelerate library for distributed training. We optimize the next-token prediction loss and select the checkpoint with the lowest validation loss. Training takes 5 epochs until convergence, with an effective batch size of 32 and AdamW optimization using a weight decay of $0.01$ and a linear learning-rate schedule that peaks at $0.0002$ after a $100$-step warm-up. LoRA is configured with rank $r = 64$ and scaling parameter $\alpha = 128$, selected via a small ablation over both hyperparameters reported in Appendix~\ref{app:lora_ablation}. The context length is limited to $6144$ during training to ensure that every training example fits within the limit, and generation is capped at $3328$. At inference we use sampling with temperature $0.7$, top-$k$ $=50$, and top-$p$ $=0.95$.
\section{On the inadequacy of FID for floor plan evaluation}
\label{app:fid_inadequacy}
We omit the Fr\'echet Inception Distance (FID), which prior raster floor plan
work~\cite{housegan++,housediffusion} reports as a ``diversity'' score, for
three reasons.

\paragraph{Construct validity.} FID measures distributional similarity in
ImageNet-trained Inception-v3 feature space and is structurally blind to the
properties our task actually requires: an exact room multiset
($\mathcal{A}_{tc}$), correct access-graph connectivity (GED, $\mathcal{A}$),
and gap-/overlap-free tiling of the input boundary ($\rho_{\mathrm{out}}$,
$\rho_{\mathrm{ovl}}$). A model can attain a low FID while violating all of
these; conversely, a model that satisfies them exactly --- as
HypergraphFormer does by construction --- gains no credit under FID.

\paragraph{Out-of-domain feature backbone.} The Inception-v3 embedding
underlying FID was calibrated on natural photographs, not top-down
color-coded room rasters. Recent work shows that FID is statistically biased
(sample-size and model dependent) and frequently disagrees with human
judgments outside its training domain, so its absolute values are not a
reliable proxy for plan quality in our setting.

\paragraph{Baselines themselves treat FID as insufficient.} The most recent
vector baseline we compare against, DiffPlanner~\cite{DiffPlanner}, states
that FID ``fails to specifically account for the intricate geometric and
topological details in the quality of generated [\dots] floor plans'' and
supplements it with the same family of geometry/topology statistics we
already report. HouseDiffusion~\cite{housediffusion} likewise withholds its
realism/FID score in the non-Manhattan regime. We therefore restrict reporting to the structural, geometric, and
tiling metrics defined in Sec.~\ref{sec:metrics}, which together constitute a
stricter and more diagnostic test of floor plan validity than FID.
\section{Editing tasks}
\label{app:editing_tasks}

This appendix complements the headline procedural-editing summary in the main paper (Section~\ref{sec:ablation_studies}) by (i) listing the procedural-editing algorithms (Section~\ref{app:editing_tasks_algorithms}), (ii) reporting full dataset-level aggregates and the corresponding ablation analysis (Section~\ref{app:editing_tasks_aggregate}, Table~\ref{tab:ablation_studies_summary}), (iii) providing per room-count bin breakdowns of each editing stage on RPLAN (Table~\ref{tab:per_bin_hf_ablation_rplan}) and WMR24 (Table~\ref{tab:per_bin_hf_ablation_wmr24}), and (iv) cataloguing the complex tool-call edits illustrated in the main paper (Section~\ref{app:complex_editing_tasks}).

\subsection{Algorithms}
\label{app:editing_tasks_algorithms}

The procedural-editing pipeline summarized in Section~\ref{sec:procedural-editing} proceeds in three stages: add/remove rooms to align with the input access graph (Algorithm~\ref{alg:addremove}), rotate or flip the hypergraph to improve compactness (Algorithm~\ref{alg:rotate}), and gradient-descent refinement of the BSP split parameters (Algorithm~\ref{alg:parametric}). We describe each stage in turn.

\subsubsection{Add/remove rooms.}
We compare the room-type multiset of the predicted hypergraph $\mathcal{H}$ against the one implied by the input access graph $\mathcal{G}_{\!\mathrm{acc}}$. Whenever the two disagree, $\mathrm{PlanOps}(\mathcal{H},\mathcal{G}_{\!\mathrm{acc}})$ emits an ordered list of add or remove operations (using the same construction rules as the dataset), which we apply in sequence via $\mathcal{H} \leftarrow \mathrm{Apply}(\mathcal{H}, c)$. After each pass, residual adjacency disagreements feed back into a fresh round of planning. The loop terminates when $\mathrm{AccessGraph}(\mathcal{H}) = \mathcal{G}_{\!\mathrm{acc}}$ or when no further operation changes the layout (Algorithm~\ref{alg:addremove}). This stage only repairs structure; it does not call the LLM.

\begin{algorithm}[H]
\caption{Procedural repair until the access graph matches the input}
\label{alg:addremove}
\begin{algorithmic}[1]
\Require hypergraph $\mathcal{H}$; input access graph $\mathcal{G}_{\!\mathrm{acc}}$
\Repeat
    \State $\mathcal{S} \gets \mathrm{PlanOps}(\mathcal{H},\mathcal{G}_{\!\mathrm{acc}})$ \Comment{multiset and/or edge gaps vs.\ $\mathcal{G}_{\!\mathrm{acc}}$}
    \If{$\mathcal{S}$ is empty}
        \State \textbf{break}
    \EndIf
    \For{$c$ in $\mathcal{S}$}
        \State $\mathcal{H} \gets \mathrm{Apply}(\mathcal{H},\, c)$
    \EndFor
\Until{$\mathrm{AccessGraph}(\mathcal{H}) = \mathcal{G}_{\!\mathrm{acc}}$ or no operation applied in the loop}
\end{algorithmic}
\end{algorithm}

\subsubsection{Rotate/mirror hypergraph.}
After multiset alignment we enumerate the eight discrete BSP-level rigid transforms (four rotations $\theta \in \{0^\circ,90^\circ,180^\circ,270^\circ\}$, with and without reflection) so that footprints stay axis-aligned and room identities are preserved. Each candidate is scored by reusing the per-polygon compactness deviation $\delta(a,b)$ of Section~\ref{sec:metrics}, but with $b$ set to a square of the same area as $a$ (which makes the score ground-truth-free and therefore usable at inference time). The candidate's score is then the mean of $\delta(a, \square_a)$ over its rooms, and we select the minimizing $(\mathrm{refl}^\star,\theta^\star)$ among the candidates whose mesh-derived access graph still equals $\mathcal{G}_{\!\mathrm{acc}}$, discarding the rest (Algorithm~\ref{alg:rotate}).

\begin{algorithm}[H]
\caption{Rotation/mirror selection by per-room square-compactness}
\label{alg:rotate}
\begin{algorithmic}[1]
\Require hypergraph $\mathcal{H}$ after multiset alignment with $n$ rooms; input access graph $\mathcal{G}_{\!\mathrm{acc}}$
\State $\mathcal{F} \gets \emptyset$ \Comment{feasible (refl, $\theta$) candidates}
\For{$(\mathrm{refl},\theta) \in \{0,1\} \times \{0^\circ,90^\circ,180^\circ,270^\circ\}$} \Comment{$\mathrm{refl}{=}0$: identity; $\mathrm{refl}{=}1$: horizontal flip}
    \State $\mathcal{H}_{\mathrm{cand}} \gets \mathrm{Apply}(\mathcal{H}, \mathrm{refl}, \theta)$ at the BSP root
    \If{$\mathrm{AccessGraph}(\mathcal{H}_{\mathrm{cand}}) = \mathcal{G}_{\!\mathrm{acc}}$}
        \State $\mathcal{F} \gets \mathcal{F} \cup \{(\mathrm{refl},\theta,\mathcal{H}_{\mathrm{cand}})\}$
    \EndIf
\EndFor
\State $(\mathrm{refl}^\star,\theta^\star,\mathcal{H}^\star) \gets \arg\min_{(\mathrm{refl},\theta,\mathcal{H}_{\mathrm{cand}}) \in \mathcal{F}} \;\tfrac{1}{n}\sum_{a \in \mathcal{H}_{\mathrm{cand}}} \delta(a,\square_a)$
\State \Return $\mathcal{H}^\star$
\end{algorithmic}
\end{algorithm}

\subsubsection{Parametric optimization.}
At every internal node of the BSP tree, the parent region is split between its two children by a single area ratio $s \in (0,1)$, where $s$ is the fraction of the parent's area assigned to the left child (and $1{-}s$ to the right). Stacking these ratios across all $d$ internal nodes yields the parameter vector $\mathbf{s} \in (0,1)^d$, which fully determines the geometric layout: varying $\mathbf{s}$ reallocates room areas without changing the BSP topology, room types, or count.

Two area objectives, both ground-truth-free at inference time, are derived from the metrics of Section~\ref{sec:metrics}.

\paragraph{Compactness.} For each room $a$ in the layout induced by $\mathbf{s}$, we reuse the per-polygon deviation $\delta(a,\square_a)$ from Algorithm~\ref{alg:rotate} and average over the $n$ rooms:
\begin{equation}
    \delta(\mathbf{s}) \;=\; \tfrac{1}{n} \sum_{a \in \mathcal{H}(\mathbf{s})} \delta\!\bigl(a, \square_a\bigr).
    \label{eq:delta_s}
\end{equation}
This is the same per-room mean square-deviation that Algorithm~\ref{alg:rotate} minimizes by enumeration; here we minimize it by gradient descent in $\mathbf{s}$.

\paragraph{Area allocation.} The test-time metric $\varepsilon(P_i,G_i)$ of~\eqref{eq:epsilon} compares predicted area proportions against ground truth, which is unavailable at inference. We replace the ground-truth proportions with a fixed canonical reference table $\boldsymbol{\pi}^{\mathrm{ref}}_{(g)}$ indexed by bedroom tier $g \in \{\text{studio}, 1\text{-bed}, 2\text{-bed}, 3^{+}\text{-bed}\}$, read off from $\mathcal{G}_{\!\mathrm{acc}}$ and reported in Table~\ref{tab:pi_ref}. The optimization-time loss is then
\begin{equation}
    \varepsilon(\mathbf{s}) \;=\; \varepsilon\!\bigl(\mathcal{H}(\mathbf{s}),\; \boldsymbol{\pi}^{\mathrm{ref}}_{(g)}\bigr),
    \label{eq:eps_s}
\end{equation}
applying the same MAE-on-normalized-proportions formula as~\eqref{eq:epsilon} but with the canonical reference in the second slot. We use the same notation $\varepsilon$ as the test-time metric, distinguishing the two by their argument shape: $\varepsilon(P_i, G_i)$ takes apartments while $\varepsilon(\mathbf{s})$ takes split parameters.

\begin{table}[h]
    \centering
    \footnotesize
    \caption{Canonical per-room area fractions $\boldsymbol{\pi}^{\mathrm{ref}}_{(g)}$ used by the area-allocation loss in~\eqref{eq:eps_s}, computed as the average per-room area fraction in the RPLAN training set, broken down by bedroom tier. Cells with no entry (\texttt{--}) indicate room types not present in that tier. Values are rounded to two decimal places.}
    \label{tab:pi_ref}
    \setlength{\tabcolsep}{6pt}
    \renewcommand{\arraystretch}{1.05}
    \begin{tabular}{@{}l c c c c c c@{}}
        \toprule
        Tier $g$ & living & kitchen & bed & bath & balcony & storage \\
        \midrule
        studio       & $0.56$ & $0.08$ & --     & $0.06$ & $0.20$ & --     \\
        $1$-bed      & $0.42$ & $0.10$ & $0.25$ & $0.08$ & $0.08$ & $0.05$ \\
        $2$-bed      & $0.42$ & $0.08$ & $0.17$ & $0.06$ & $0.06$ & $0.04$ \\
        $3^{+}$-bed  & $0.40$ & $0.07$ & $0.14$ & $0.05$ & $0.06$ & $0.03$ \\
        \bottomrule
    \end{tabular}
\end{table}

\paragraph{Optimization modes.} We support three modes: \emph{(i)}~\emph{$\delta$ only:} minimize $\delta(\mathbf{s})$. \emph{(ii)}~\emph{$\varepsilon$ only:} minimize $\varepsilon(\mathbf{s})$ and \emph{(iii)}~\emph{$\delta{+}\varepsilon$:} given initial splits $\mathbf{s}_0$, fix the weight $w = \varepsilon(\mathbf{s}_0)/\max\!\bigl(\eta,\,\delta(\mathbf{s}_0)\bigr)$ for some floor $\eta > 0$ and minimize the combined loss
\begin{equation}
  \mathcal{L}_{\mathrm{combined}}(\mathbf{s})
  \;=\;
  w\,\delta(\mathbf{s})
  \;+\;
  \varepsilon(\mathbf{s}).
  \label{eq:combined}
\end{equation}
The weight $w$ rebalances the two losses so that, at $\mathbf{s}_0$, their contributions are comparable in magnitude.

\paragraph{Access-graph penalty.} Across all three modes, every loss evaluation also incurs an additive penalty $\lambda \cdot \mathbf{1}\bigl[\mathrm{AccessGraph}(\mathcal{H}(\mathbf{s})) \neq \mathcal{G}_{\!\mathrm{acc}}\bigr]$ whenever the perturbed layout's access graph disagrees with the input (we use $\lambda = 0.05$, comparable to the mean values of $\delta$ and $\varepsilon$). We optimize via projected gradient descent until convergence or an iteration cap (Algorithm~\ref{alg:parametric}).

\begin{algorithm}[H]
\caption{Parametric area-split refinement}
\label{alg:parametric}
\begin{algorithmic}[1]
\Require hypergraph $\mathcal{H}_0$ after rotation, with $d$ internal BSP nodes and $n$ rooms; bedroom tier $g$; objective $\in \{\delta,\;\varepsilon,\;\delta{+}\varepsilon\}$; reference table $\boldsymbol{\pi}^{\mathrm{ref}}_{(g)}$; access graph $\mathcal{G}_{\!\mathrm{acc}}$
\State $\mathbf{s}_0 \gets$ collect area-split ratios from $\mathcal{H}_0$
\If{objective $= \delta{+}\varepsilon$}
    \State $w \gets \varepsilon(\mathbf{s}_0) \,\big/\, \max\!\bigl(\eta,\,\delta(\mathbf{s}_0)\bigr)$
\EndIf
\State define $\mathcal{L}(\mathbf{s})$ as $\delta(\mathbf{s})$, $\varepsilon(\mathbf{s})$, or $\mathcal{L}_{\mathrm{combined}}(\mathbf{s})$ per the chosen objective
\State $\mathbf{s} \gets \mathbf{s}_0$
\Repeat
    \State $\mathcal{L}_{\mathrm{tot}}(\mathbf{s}) \gets \mathcal{L}(\mathbf{s}) + \lambda \cdot \mathbf{1}\bigl[\mathrm{AccessGraph}(\mathcal{H}(\mathbf{s})) \neq \mathcal{G}_{\!\mathrm{acc}}\bigr]$
    \State $\mathbf{s} \gets \mathbf{s} - \nabla \mathcal{L}_{\mathrm{tot}}(\mathbf{s})$ \Comment{projected gradient step in $(0,1)^d$}
\Until{termination criterion}
\State \Return $\mathcal{H}(\mathbf{s})$
\end{algorithmic}
\end{algorithm}

\subsection{Aggregate ablations of the procedural edits}
\label{app:editing_tasks_aggregate}
This subsection complements the headline summary in Section~\ref{sec:ablation_studies} (main paper) by reporting dataset-level aggregates for each stage of the procedural-editing pipeline and discussing the contribution of each stage in detail.

\begin{table}[h]
    \centering
    \footnotesize
    \caption{HypergraphFormer post-processing ablations on \textbf{RPLAN} (top) and \textbf{WMR24} (bottom), reporting dataset-level aggregates only. All rows report our method; rows differ only in which post-processing stage is applied (\emph{Raw LLM Output} is HypergraphFormer's untouched output). $\mathcal{A}$ is the GED accuracy ($\%$ of samples with $\mathrm{GED}=0$); $\mathcal{A}_{tc}$ is the multiset accuracy of Section~\ref{sec:metrics}. Per-bin breakdowns are reported in Tables~\ref{tab:per_bin_hf_ablation_rplan} and~\ref{tab:per_bin_hf_ablation_wmr24}.}
    \label{tab:ablation_studies_summary}
    \resizebox{\columnwidth}{!}{%
    \setlength{\tabcolsep}{5pt}
    \renewcommand{\arraystretch}{1.12}
    \begin{tabular}{@{} >{\raggedright\arraybackslash}p{4.15cm} *{5}{c} @{}}
        \multicolumn{6}{@{}l}{\textbf{RPLAN}} \\
        \toprule
        Variant
        & \shortstack{GED\\{\scriptsize$(\downarrow)$}}
        & \shortstack{$\mathcal{A}$\\{\scriptsize$(\uparrow)$}}
        & \shortstack{$\mathcal{A}_{tc}$\\{\scriptsize$(\uparrow)$}}
        & \shortstack{$\delta$\\{\scriptsize$(\downarrow)$}}
        & \shortstack{$\varepsilon$ (\%)\\{\scriptsize$(\downarrow)$}} \\
        \midrule
        Raw LLM Output
            & $1.99 {\scriptscriptstyle \pm 0.02}$ & $35.5 {\scriptscriptstyle \pm 0.29}$ & $75.4 {\scriptscriptstyle \pm 0.38}$ & $0.142 {\scriptscriptstyle \pm 0.001}$ & $3.96 {\scriptscriptstyle \pm 0.01}$ \\
        Add/Remove Rooms
            & $1.72 {\scriptscriptstyle \pm 0.02}$ & $39.1 {\scriptscriptstyle \pm 0.23}$ & $100.0 {\scriptscriptstyle \pm 0.00}$ & $0.146 {\scriptscriptstyle \pm 0.001}$ & $3.94 {\scriptscriptstyle \pm 0.01}$ \\
        Pick Orientation
            & $1.63 {\scriptscriptstyle \pm 0.02}$ & $40.6 {\scriptscriptstyle \pm 0.11}$ & $100.0 {\scriptscriptstyle \pm 0.00}$ & $0.103 {\scriptscriptstyle \pm 0.000}$ & $3.94 {\scriptscriptstyle \pm 0.01}$ \\
        Optimize $\varepsilon$
            & $1.60 {\scriptscriptstyle \pm 0.02}$ & $41.3 {\scriptscriptstyle \pm 0.15}$ & $100.0 {\scriptscriptstyle \pm 0.00}$ & $0.117 {\scriptscriptstyle \pm 0.001}$ & $3.00 {\scriptscriptstyle \pm 0.00}$ \\
        Optimize $\delta$ and $\varepsilon$
            & $1.62 {\scriptscriptstyle \pm 0.02}$ & $40.9 {\scriptscriptstyle \pm 0.12}$ & $100.0 {\scriptscriptstyle \pm 0.00}$ & $0.0986 {\scriptscriptstyle \pm 0.0002}$ & $3.37 {\scriptscriptstyle \pm 0.01}$ \\
        \addlinespace[0.4em]
        \multicolumn{6}{@{}l}{\textbf{WMR24}} \\
        \midrule
        Variant
        & \shortstack{GED\\{\scriptsize$(\downarrow)$}}
        & \shortstack{$\mathcal{A}$\\{\scriptsize$(\uparrow)$}}
        & \shortstack{$\mathcal{A}_{tc}$\\{\scriptsize$(\uparrow)$}}
        & \shortstack{$\delta$\\{\scriptsize$(\downarrow)$}}
        & \shortstack{$\varepsilon$ (\%)\\{\scriptsize$(\downarrow)$}} \\
        \midrule
        Raw LLM Output
            & $1.99 {\scriptscriptstyle \pm 0.03}$ & $48.0 {\scriptscriptstyle \pm 0.8}$ & $75.4 {\scriptscriptstyle \pm 1.0}$ & $0.131 {\scriptscriptstyle \pm 0.001}$ & $6.34 {\scriptscriptstyle \pm 0.10}$ \\
        Add/Remove Rooms
            & $1.73 {\scriptscriptstyle \pm 0.05}$ & $52.1 {\scriptscriptstyle \pm 0.9}$ & $99.9 {\scriptscriptstyle \pm 0.1}$ & $0.135 {\scriptscriptstyle \pm 0.001}$ & $6.28 {\scriptscriptstyle \pm 0.09}$ \\
        Pick Orientation
            & $1.68 {\scriptscriptstyle \pm 0.04}$ & $52.3 {\scriptscriptstyle \pm 0.7}$ & $99.9 {\scriptscriptstyle \pm 0.1}$ & $0.097 {\scriptscriptstyle \pm 0.001}$ & $6.28 {\scriptscriptstyle \pm 0.09}$ \\
        Optimize $\varepsilon$
            & $1.75 {\scriptscriptstyle \pm 0.04}$ & $51.2 {\scriptscriptstyle \pm 0.8}$ & $99.9 {\scriptscriptstyle \pm 0.1}$ & $0.134 {\scriptscriptstyle \pm 0.001}$ & $4.74 {\scriptscriptstyle \pm 0.04}$ \\
        Optimize $\delta$ and $\varepsilon$
            & $1.72 {\scriptscriptstyle \pm 0.03}$ & $51.6 {\scriptscriptstyle \pm 0.8}$ & $99.9 {\scriptscriptstyle \pm 0.1}$ & $0.070 {\scriptscriptstyle \pm 0.001}$ & $5.45 {\scriptscriptstyle \pm 0.07}$ \\
        \bottomrule
    \end{tabular}%
    }
\end{table}

We isolate the contribution of each component of the post-processing pipeline by starting from the raw LLM output and applying the editing stages in sequence. Table~\ref{tab:ablation_studies_summary} reports dataset-level aggregates on RPLAN (top) and WMR24 (bottom). The first three rows compose: \emph{Add/Remove Rooms} is applied to the raw output, and \emph{Pick Orientation} is applied to the add/remove result. The two \emph{Optimize} rows are alternative branches: each starts from \emph{Pick Orientation} and runs gradient-descent optimization with a different objective.

\emph{Add/Remove Rooms} is the single largest jump in the pipeline. It raises $\mathcal{A}_{tc}$ from $\approx\!75\%$ to $100\%$ on RPLAN and to $99.9\%$ on WMR24, since by construction the operation forces the predicted room multiset to match the input access graph (Algorithm~\ref{alg:addremove}). It also pulls GED down ($1.99 \to 1.72$ on RPLAN, $1.99 \to 1.72$ on WMR24) because correcting room counts removes the dominant class of structural errors. \emph{Pick Orientation} then rotates and reflects the layout to maximize per-room compactness, reducing $\delta$ by roughly $25\%$ on both datasets ($0.142 \to 0.103$ on RPLAN, $0.131 \to 0.097$ on WMR24) with a small additional GED gain. The two parametric branches then trade off $\delta$ against $\varepsilon$: optimizing $\varepsilon$ alone drives the area error down (RPLAN: $3.94\% \to 3.00\%$; WMR24: $6.28\% \to 4.74\%$) but lifts $\delta$, since shrinking large rooms toward their reference proportions tends to elongate them; jointly optimizing $\delta$ and $\varepsilon$ recovers the lowest $\delta$ on both datasets ($0.0986$ on RPLAN, $0.070$ on WMR24) at a modest cost in $\varepsilon$. Strict GED accuracy $\mathcal{A}$ is essentially unaffected by the geometric optimizers ($\approx\!41\%$ on RPLAN, $\approx\!52\%$ on WMR24), confirming that they refine geometry without altering the underlying graph structure already fixed by \emph{Add/Remove Rooms} and \emph{Pick Orientation}.

The pipeline order matters: each stage assumes the previous one has stabilized the graph or geometry it depends on. \emph{Add/Remove Rooms} must finalize the room multiset before \emph{Pick Orientation} chooses an orientation, and the parametric optimizers must run on a fixed multiset and orientation, since their gradients act on per-node area splits in the BSP tree. Within this order, however, the geometric stages can be selected per use case: jointly optimizing $\delta$ and $\varepsilon$ is the most balanced default, optimizing $\varepsilon$ alone is preferable when area fidelity is paramount, and skipping the optimizers retains the strict graph-level accuracy at the lowest computational cost.

\subsection{Per-room detailed metrics}
\label{app:editing_tasks_per_room}

This subsection breaks down the aggregate ablations of Section~\ref{app:editing_tasks_aggregate} by room-count bin (same bins as Table~\ref{tab:per_bin_comparison}), separately for RPLAN (Table~\ref{tab:per_bin_hf_ablation_rplan}) and WMR24 (Table~\ref{tab:per_bin_hf_ablation_wmr24}). For each editing stage, we report GED, the GED accuracy $\mathcal{A}$, the multiset accuracy $\mathcal{A}_{tc}$, and the geometric metrics $\delta$ and $\varepsilon$, allowing one to verify that the per-stage trends discussed in the aggregate are stable across room-count complexity.

\begin{table*}[h]
    \centering
    \footnotesize
    \caption{Per-bin comparison of HypergraphFormer post-processing ablations on \textbf{RPLAN} (same metrics and bins as Table~\ref{tab:per_bin_comparison}). All rows report our method; rows differ only in which post-processing stage is applied (\emph{Raw LLM Output} is HypergraphFormer's untouched output). $\mathcal{A}$ reports the \emph{GED accuracy} (fraction of samples with GED${}=0$, in \%); $\mathcal{A}_{tc}$ follows the multiset accuracy in Section~\ref{sec:metrics}.}
    \renewcommand{\arraystretch}{1.05}
    \begin{tabular}{l l c c c c c c c}
        \toprule
        Metric & Variant & Agg. & $\leq 4$ & 5 & 6 & 7 & 8 & $9 \leq$ \\
        \midrule
        \multirow{5}{*}{GED ($\downarrow$)}
        & Raw LLM Output & $1.99 {\scriptscriptstyle \pm 0.02}$ & $1.92$ & $1.88$ & $1.94$ & $2.02$ & $2.06$ & N/A \\
        & Add/Remove Rooms & $1.72 {\scriptscriptstyle \pm 0.02}$ & $1.46$ & $1.55$ & $1.64$ & $1.75$ & $1.82$ & N/A \\
        & Pick Orientation & $1.63 {\scriptscriptstyle \pm 0.02}$ & $1.54$ & $1.53$ & $1.57$ & $1.66$ & $1.71$ & N/A \\
        & Optimize $\varepsilon$ & $1.60 {\scriptscriptstyle \pm 0.02}$ & $1.43$ & $1.47$ & $1.54$ & $1.62$ & $1.69$ & N/A \\
        & Optimize $\delta$ and $\varepsilon$ & $1.62 {\scriptscriptstyle \pm 0.02}$ & $1.50$ & $1.45$ & $1.58$ & $1.63$ & $1.71$ & N/A \\
        \cmidrule(lr){1-9}
        \multirow{5}{*}{$\mathcal{A}$ ($\uparrow$)}
        & Raw LLM Output & $35.5 {\scriptscriptstyle \pm 0.29}$ & $42.3$ & $38.4$ & $36.2$ & $35.2$ & $34.1$ & N/A \\
        & Add/Remove Rooms & $39.1 {\scriptscriptstyle \pm 0.23}$ & $42.9$ & $42.8$ & $40.8$ & $38.2$ & $37.1$ & N/A \\
        & Pick Orientation & $40.6 {\scriptscriptstyle \pm 0.11}$ & $42.9$ & $43.5$ & $41.7$ & $40.0$ & $39.0$ & N/A \\
        & Optimize $\varepsilon$ & $41.3 {\scriptscriptstyle \pm 0.15}$ & $46.9$ & $43.2$ & $43.1$ & $40.6$ & $39.3$ & N/A \\
        & Optimize $\delta$ and $\varepsilon$ & $40.9 {\scriptscriptstyle \pm 0.12}$ & $46.3$ & $45.8$ & $41.7$ & $40.6$ & $38.8$ & N/A \\
        \cmidrule(lr){1-9}
        \multirow{5}{*}{$\mathcal{A}_{tc}$ ($\uparrow$)}
        & Raw LLM Output & $75.4 {\scriptscriptstyle \pm 0.38}$ & $80.0$ & $85.5$ & $82.3$ & $74.0$ & $66.0$ & N/A \\
        & Add/Remove Rooms & $100.0 {\scriptscriptstyle \pm 0.00}$ & $100.0$ & $100.0$ & $100.0$ & $100.0$ & $100.0$ & N/A \\
        & Pick Orientation & $100.0 {\scriptscriptstyle \pm 0.00}$ & $100.0$ & $100.0$ & $100.0$ & $100.0$ & $100.0$ & N/A \\
        & Optimize $\varepsilon$ & $100.0 {\scriptscriptstyle \pm 0.00}$ & $100.0$ & $100.0$ & $100.0$ & $100.0$ & $100.0$ & N/A \\
        & Optimize $\delta$ and $\varepsilon$ & $100.0 {\scriptscriptstyle \pm 0.00}$ & $100.0$ & $100.0$ & $100.0$ & $100.0$ & $100.0$ & N/A \\
        \cmidrule(lr){1-9}
        \multirow{5}{*}{$\delta$ ($\downarrow$)}
        & Raw LLM Output & $0.142 {\scriptscriptstyle \pm 0.001}$ & $0.137$ & $0.124$ & $0.133$ & $0.144$ & $0.154$ & N/A \\
        & Add/Remove Rooms & $0.146 {\scriptscriptstyle \pm 0.001}$ & $0.138$ & $0.127$ & $0.135$ & $0.148$ & $0.161$ & N/A \\
        & Pick Orientation & $0.103 {\scriptscriptstyle \pm 0.000}$ & $0.082$ & $0.082$ & $0.093$ & $0.106$ & $0.118$ & N/A \\
        & Optimize $\varepsilon$ & $0.117 {\scriptscriptstyle \pm 0.001}$ & $0.094$ & $0.095$ & $0.107$ & $0.119$ & $0.131$ & N/A \\
        & Optimize $\delta$ and $\varepsilon$ & $0.0986 {\scriptscriptstyle \pm 0.0002}$ & $0.081$ & $0.081$ & $0.092$ & $0.100$ & $0.109$ & N/A \\
        \cmidrule(lr){1-9}
        \multirow{5}{*}{$\varepsilon$ (\%, $\downarrow$)}
        & Raw LLM Output & $3.96 {\scriptscriptstyle \pm 0.01}$ & $6.5$ & $4.8$ & $4.1$ & $3.9$ & $3.7$ & N/A \\
        & Add/Remove Rooms & $3.94 {\scriptscriptstyle \pm 0.01}$ & $6.3$ & $4.7$ & $4.1$ & $3.8$ & $3.7$ & N/A \\
        & Pick Orientation & $3.94 {\scriptscriptstyle \pm 0.01}$ & $6.4$ & $4.7$ & $4.1$ & $3.8$ & $3.7$ & N/A \\
        & Optimize $\varepsilon$ & $3.00 {\scriptscriptstyle \pm 0.00}$ & $4.4$ & $3.5$ & $3.1$ & $2.9$ & $2.8$ & N/A \\
        & Optimize $\delta$ and $\varepsilon$ & $3.37 {\scriptscriptstyle \pm 0.01}$ & $5.2$ & $3.8$ & $3.5$ & $3.3$ & $3.2$ & N/A \\
        \bottomrule
    \end{tabular}
    \label{tab:per_bin_hf_ablation_rplan}
\end{table*}

\begin{table*}[h]
    \centering
    \footnotesize
    \caption{Per-bin comparison of HypergraphFormer post-processing ablations on \textbf{WMR24} (same layout as Table~\ref{tab:per_bin_hf_ablation_rplan}). All rows report our method; rows differ only in which post-processing stage is applied (\emph{Raw LLM Output} is HypergraphFormer's untouched output).}
    \renewcommand{\arraystretch}{1.05}
    \begin{tabular}{l l c c c c c c c}
        \toprule
        Metric & Variant & Agg. & $\leq 4$ & 5 & 6 & 7 & 8 & $9 \leq$ \\
        \midrule
        \multirow{5}{*}{GED ($\downarrow$)}
        & Raw LLM Output & $1.99 {\scriptscriptstyle \pm 0.03}$ & $1.96$ & $2.14$ & $1.94$ & $1.94$ & $1.79$ & $2.03$ \\
        & Add/Remove Rooms & $1.73 {\scriptscriptstyle \pm 0.05}$ & $1.32$ & $1.52$ & $1.58$ & $1.64$ & $1.93$ & $1.96$ \\
        & Pick Orientation & $1.68 {\scriptscriptstyle \pm 0.04}$ & $1.44$ & $1.61$ & $1.52$ & $1.68$ & $1.75$ & $1.83$ \\
        & Optimize $\varepsilon$ & $1.75 {\scriptscriptstyle \pm 0.04}$ & $1.39$ & $1.50$ & $1.64$ & $1.60$ & $1.78$ & $2.03$ \\
        & Optimize $\delta$ and $\varepsilon$ & $1.72 {\scriptscriptstyle \pm 0.03}$ & $1.44$ & $1.60$ & $1.60$ & $1.68$ & $1.91$ & $1.86$ \\
        \cmidrule(lr){1-9}
        \multirow{5}{*}{$\mathcal{A}$ ($\uparrow$)}
        & Raw LLM Output & $48.0 {\scriptscriptstyle \pm 0.8}$ & $50.2$ & $44.2$ & $48.6$ & $46.9$ & $53.0$ & $47.7$ \\
        & Add/Remove Rooms & $52.1 {\scriptscriptstyle \pm 0.9}$ & $59.5$ & $54.0$ & $53.5$ & $54.2$ & $49.7$ & $48.6$ \\
        & Pick Orientation & $52.3 {\scriptscriptstyle \pm 0.7}$ & $55.5$ & $53.4$ & $56.4$ & $51.5$ & $50.9$ & $49.9$ \\
        & Optimize $\varepsilon$ & $51.2 {\scriptscriptstyle \pm 0.8}$ & $58.0$ & $57.0$ & $52.6$ & $55.7$ & $50.0$ & $45.0$ \\
        & Optimize $\delta$ and $\varepsilon$ & $51.6 {\scriptscriptstyle \pm 0.8}$ & $56.0$ & $52.2$ & $53.5$ & $51.3$ & $50.1$ & $49.8$ \\
        \cmidrule(lr){1-9}
        \multirow{5}{*}{$\mathcal{A}_{tc}$ ($\uparrow$)}
        & Raw LLM Output & $74.0 {\scriptscriptstyle \pm 0.9}$ & $81.1$ & $86.5$ & $85.9$ & $80.5$ & $73.5$ & $59.9$ \\
        & Add/Remove Rooms & $99.9 {\scriptscriptstyle \pm 0.1}$ & $100.0$ & $100.0$ & $100.0$ & $100.0$ & $100.0$ & $99.8$ \\
        & Pick Orientation & $99.9 {\scriptscriptstyle \pm 0.1}$ & $100.0$ & $100.0$ & $100.0$ & $100.0$ & $100.0$ & $99.8$ \\
        & Optimize $\varepsilon$ & $99.9 {\scriptscriptstyle \pm 0.1}$ & $100.0$ & $100.0$ & $100.0$ & $100.0$ & $100.0$ & $99.8$ \\
        & Optimize $\delta$ and $\varepsilon$ & $99.9 {\scriptscriptstyle \pm 0.1}$ & $100.0$ & $100.0$ & $100.0$ & $100.0$ & $100.0$ & $99.8$ \\
        \cmidrule(lr){1-9}
        \multirow{5}{*}{$\delta$ ($\downarrow$)}
        & Raw LLM Output & $0.131 {\scriptscriptstyle \pm 0.001}$ & $0.107$ & $0.123$ & $0.114$ & $0.131$ & $0.130$ & $0.149$ \\
        & Add/Remove Rooms & $0.135 {\scriptscriptstyle \pm 0.001}$ & $0.106$ & $0.123$ & $0.117$ & $0.133$ & $0.133$ & $0.155$ \\
        & Pick Orientation & $0.097 {\scriptscriptstyle \pm 0.001}$ & $0.073$ & $0.077$ & $0.083$ & $0.094$ & $0.099$ & $0.117$ \\
        & Optimize $\varepsilon$ & $0.134 {\scriptscriptstyle \pm 0.001}$ & $0.095$ & $0.110$ & $0.122$ & $0.134$ & $0.142$ & $0.157$ \\
        & Optimize $\delta$ and $\varepsilon$ & $0.070 {\scriptscriptstyle \pm 0.001}$ & $0.063$ & $0.059$ & $0.061$ & $0.069$ & $0.069$ & $0.079$ \\
        \cmidrule(lr){1-9}
        \multirow{5}{*}{$\varepsilon$ (\%, $\downarrow$)}
        & Raw LLM Output & $6.34 {\scriptscriptstyle \pm 0.10}$ & $9.65$ & $8.35$ & $7.27$ & $6.43$ & $5.74$ & $4.39$ \\
        & Add/Remove Rooms & $6.28 {\scriptscriptstyle \pm 0.09}$ & $9.45$ & $8.28$ & $7.18$ & $6.37$ & $5.63$ & $4.40$ \\
        & Pick Orientation & $6.28 {\scriptscriptstyle \pm 0.09}$ & $9.44$ & $8.28$ & $7.17$ & $6.37$ & $5.63$ & $4.39$ \\
        & Optimize $\varepsilon$ & $4.74 {\scriptscriptstyle \pm 0.04}$ & $7.68$ & $6.37$ & $5.59$ & $4.70$ & $4.09$ & $3.13$ \\
        & Optimize $\delta$ and $\varepsilon$ & $5.45 {\scriptscriptstyle \pm 0.07}$ & $7.38$ & $6.52$ & $6.34$ & $5.63$ & $5.18$ & $4.14$ \\
        \bottomrule
    \end{tabular}
    \label{tab:per_bin_hf_ablation_wmr24}
\end{table*}

\subsection{LLM tool calls for complex edits}
\label{app:complex_editing_tasks}
This subsection complements Section~\ref{sec:complex_editing_tasks} (main paper) by enumerating the complex edits illustrated in Fig.~\ref{fig:complex_editing_tasks_examples} and listing a sample tool call for each. All edits operate directly on the BSP tree and access graph and require no further LLM generation beyond parsing the tool call.

\noindent\textit{Remove Room.}\quad
Building on the repair logic of Algorithm~\ref{alg:addremove}, the user specifies a room to delete by name. A simple traversal removes the corresponding BSP leaf, and the freed area is redistributed among neighboring rooms by propagating the change up the tree, so that no single sibling absorbs a disproportionate share. Sample tool call: \texttt{delete room <room\_name>}.

\noindent\textit{Add Room.}\quad
Symmetric to the previous edit: the user specifies a new room together with an existing room to attach it to. A new BSP leaf is inserted alongside the reference leaf, and the area required for the new room is drawn from neighboring rooms, with the change propagated up the tree to spread the contraction rather than concentrate it on a single neighbor. Sample tool call: \texttt{add room <new\_room> next to <reference\_room>}.

\noindent\textit{Resize Room.}\quad
Scales a target room by a user-specified factor and scales the surrounding rooms inversely so that the scaling effect propagates evenly through the BSP tree, preserving the apartment's outer boundary. Sample tool call: \texttt{resize <room\_name> by <factor>}.

\noindent\textit{Rotate Layout.}\quad
Rotates the entire hypergraph by a user-specified angle, leaving room semantics unchanged. Sample tool call: \texttt{rotate layout by <angle>}.

\noindent\textit{Move Entrance.}\quad
A variant of \textit{Rotate Layout} that orients the apartment so the entrance faces a chosen direction. The command identifies the room connected to the \emph{Outside} node, determines which side of the boundary it currently abuts (left, right, top, or bottom), and applies the rotation that aligns it with the requested direction. Sample tool call: \texttt{move door to <direction>}.

\noindent\textit{Orient Specific Room.}\quad
Generalizes \textit{Move Entrance} to any room type: the layout is rotated so that a designated room (e.g., a bedroom) is positioned along a requested side of the apartment. This is useful, for example, for orienting facade-facing rooms toward the well-lit side of the building. Sample tool call: \texttt{orient <room\_type> to <direction>}.

\noindent\textit{Optimize.}\quad
Adjusts the room partitions via gradient descent to optimize a chosen geometric objective; Algorithm~\ref{alg:parametric} describes the cases of $\delta$ and $\varepsilon$ used in our post-processing pipeline. The same machinery extends naturally to other differentiable objectives, such as daylight, wind exposure, or furniture-placement scores.

\noindent\textit{Freeze and Edit.}\quad
Because the hypergraph representation is boundary-independent, edits can be applied to a subregion of an apartment or composed across multiple apartments. We illustrate this by holding one part of the floor plan fixed while regenerating the rest, allowing the user to iterate on a specific zone without disturbing the surrounding layout.

\section{Per-room detailed metrics}
\label{app:per_bin_results}

This appendix complements the main-paper aggregate comparisons (Section~\ref{sec:results}, Table~\ref{tab:input_grouped_comparison}) and the data-efficiency study (Section~\ref{sec:data_efficiency}) with per room-count breakdowns on both axes. Table~\ref{tab:per_bin_comparison} expands the access-graph block of Table~\ref{tab:input_grouped_comparison} with per-bin GED, GED accuracy $\mathcal{A}$, and the multiset accuracy $\mathcal{A}_{tc}$ for HouseGAN++ (HG), HouseDiffusion (HD), and HypergraphFormer (HF) on RPLAN and WMR24. Table~\ref{tab:per_bin_comparison_boundary} mirrors the same breakdown for the boundary-constrained block of Table~\ref{tab:input_grouped_comparison}, reporting $\mathcal{A}_{tc}$, $\delta$, $\varepsilon$, $\rho_{\mathrm{out}}$, and $\rho_{\mathrm{ovl}}$ for iPLAN, DiffPlanner (DP), and HF. Table~\ref{tab:per_bin_rplan_data_scaling} then reports the same per-bin GED and $\mathcal{A}$ for HypergraphFormer trained on progressively smaller subsets of RPLAN.

\begin{table*}[h]
    \centering
    \footnotesize
    \caption{Per-bin breakdown of the access-graph block of Table~\ref{tab:input_grouped_comparison}: HypergraphFormer (HF) against HouseGAN++ (HG) and HouseDiffusion (HD) on RPLAN and WMR24, reported as dataset aggregates (Agg.) and per room-count bins ($\leq 4$ to $9 \leq$). Metrics are GED~\eqref{eq:ged}, GED accuracy $\mathcal{A}$ (\%), and the joint type-and-count multiset accuracy $\mathcal{A}_{tc}$~\eqref{eq:accuracy}.}
    \renewcommand{\arraystretch}{1.1}
    \setlength{\tabcolsep}{4pt}
    \begin{tabular}{@{}c l l c c c c c c c@{}}
        \toprule
        Dataset & Metric & Model & Agg. & $\leq 4$ & 5 & 6 & 7 & 8 & $9 \leq$ \\
        \midrule
        \multirow{9}{*}{\rotatebox[origin=c]{90}{\makebox[0pt]{RPLAN}}}
        & \multirow{3}{*}{GED ($\downarrow$)}
        & HG & $2.59 {\scriptscriptstyle \pm 0.01}$ & $1.58$ & $1.79$ & $2.21$ & $2.67$ & $3.19$ & N/A \\
        & & HD & $1.95 {\scriptscriptstyle \pm 0.01}$ & $1.56$ & $1.72$ & $1.87$ & $2.14$ & $2.45$ & N/A \\
        & & Ours & $1.62 {\scriptscriptstyle \pm 0.02}$ & $1.50$ & $1.45$ & $1.58$ & $1.63$ & $1.71$ & N/A \\
        \cmidrule(lr){2-10}
        & \multirow{3}{*}{$\mathcal{A}$ ($\uparrow$)}
        & HG & $6.0 {\scriptscriptstyle \pm 0.12}$ & $16.3$ & $12.9$ & $8.4$ & $4.8$ & $2.6$ & N/A \\
        & & HD & $16.3 {\scriptscriptstyle \pm 0.32}$ & $23.0$ & $19.7$ & $16.8$ & $13.4$ & $10.6$ & N/A \\
        & & Ours & $40.9 {\scriptscriptstyle \pm 0.1}$ & $46.3$ & $45.8$ & $41.7$ & $40.6$ & $38.8$ & N/A \\
        \cmidrule(lr){2-10}
        & \multirow{3}{*}{$\mathcal{A}_{tc}$ ($\uparrow$)}
        & HG & $44.2 {\scriptscriptstyle \pm 0.45}$ & $62.5$ & $55.6$ & $48.4$ & $43.4$ & $36.7$ & N/A \\
        & & HD & $96.7 {\scriptscriptstyle \pm 0.04}$ & $97.9$ & $97.7$ & $97.1$ & $96.2$ & $93.6$ & N/A \\
        & & Ours & $100.0 {\scriptscriptstyle \pm 0.0}$ & $100.0$ & $100.0$ & $100.0$ & $100.0$ & $100.0$ & N/A \\
        \midrule
        \multirow{9}{*}{\rotatebox[origin=c]{90}{\makebox[0pt]{Out of Distribution (WMR24)}}}
        & \multirow{3}{*}{GED ($\downarrow$)}
        & HG & $3.80 {\scriptscriptstyle \pm 0.06}$ & $1.25$ & $1.58$ & $2.08$ & $2.78$ & $3.43$ & $6.48$ \\
        & & HD & $3.78 {\scriptscriptstyle \pm 0.03}$ & $1.89$ & $2.74$ & $3.42$ & $4.26$ & $4.89$ & $5.53$ \\
        & & Ours & $1.72 {\scriptscriptstyle \pm 0.03}$ & $1.44$ & $1.60$ & $1.60$ & $1.68$ & $1.91$ & $1.86$ \\
        \cmidrule(lr){2-10}
        & \multirow{3}{*}{$\mathcal{A}$ ($\uparrow$)}
        & HG & $8.5 {\scriptscriptstyle \pm 0.56}$ & $24.0$ & $19.8$ & $12.9$ & $6.3$ & $2.4$ & $0.6$ \\
        & & HD & $2.6 {\scriptscriptstyle \pm 0.74}$ & $11.9$ & $3.4$ & $1.6$ & $0.4$ & $0.0$ & $0.0$ \\
        & & Ours & $51.6 {\scriptscriptstyle \pm 0.8}$ & $56.0$ & $52.2$ & $53.5$ & $51.3$ & $50.1$ & $49.8$ \\
        \cmidrule(lr){2-10}
        & \multirow{3}{*}{$\mathcal{A}_{tc}$ ($\uparrow$)}
        & HG & $37.1 {\scriptscriptstyle \pm 1.02}$ & $61.7$ & $52.9$ & $49.4$ & $43.0$ & $33.7$ & $19.0$ \\
        & & HD & $80.0 {\scriptscriptstyle \pm 0.92}$ & $94.0$ & $92.5$ & $86.8$ & $80.0$ & $69.8$ & $54.9$ \\
        & & Ours & $99.9 {\scriptscriptstyle \pm 0.1}$ & $100.0$ & $100.0$ & $100.0$ & $100.0$ & $100.0$ & $99.8$ \\
        \bottomrule
    \end{tabular}
    \label{tab:per_bin_comparison}
\end{table*}

The per-bin numbers in Table~\ref{tab:per_bin_comparison} sharpen the aggregate trends discussed in Section~\ref{sec:results}. On WMR24, HG's and HD's GED grow steadily from $1.25$/$1.89$ at $\leq\!4$ rooms to $6.48$/$5.53$ at $\geq\!9$ rooms, and their $\mathcal{A}$ collapses to $0.6\%$/$0.0\%$ in the largest bin; HypergraphFormer's GED stays in a narrow band ($1.44$ to $1.91$) and its $\mathcal{A}$ stays above $49\%$ at $\geq\!9$ rooms, the only method that produces any exact matches at all in this regime. The same flatness is visible on RPLAN, where HF's per-bin GED varies by less than $0.25$ across the $\leq\!4$ to $8$-room bins. HF's $\mathcal{A}_{tc}$ stays at $100.0\%$ on every RPLAN bin and at $\geq 99.8\%$ on every WMR24 bin, confirming that the procedural add/remove edit (Algorithm~\ref{alg:addremove}) enforces the multiset constraint uniformly across complexity levels rather than only on average. Together, these breakdowns show that HypergraphFormer not only attains the strongest aggregate metrics in Table~\ref{tab:input_grouped_comparison} but also degrades gracefully along the room-count axis on which the rasterized and vectorized baselines lose ground sharply.

\begin{table*}[h]
    \centering
    \footnotesize
    \caption{Per-bin breakdown of the boundary-constrained block of Table~\ref{tab:input_grouped_comparison}: HypergraphFormer (HF) against iPLAN (IP) and DiffPlanner (DP) on RPLAN and WMR24, reported as dataset aggregates (Agg.) and per room-count bins ($\leq 4$ to $9 \leq$). Metrics are the multiset accuracy $\mathcal{A}_{tc}$~\eqref{eq:accuracy} (\%), the geometric compactness deviation $\delta$~\eqref{eq:delta}, the area proportion error $\varepsilon$~\eqref{eq:epsilon} (\%), and the boundary-normalized tiling errors $\rho_{\mathrm{out}}$ and $\rho_{\mathrm{ovl}}$~\eqref{eq:tiling} (\%). HF $\rho_{\mathrm{out}} = \rho_{\mathrm{ovl}} = 0$ on every bin by construction since BSP-derived rooms tile the apartment exactly with no gaps or overlaps.}
    \renewcommand{\arraystretch}{1.1}
    \setlength{\tabcolsep}{4pt}
    \begin{tabular}{@{}c l l c c c c c c c@{}}
        \toprule
        Dataset & Metric & Model & Agg. & $\leq 4$ & 5 & 6 & 7 & 8 & $9 \leq$ \\
        \midrule
        \multirow{15}{*}{\rotatebox[origin=c]{90}{\makebox[0pt]{RPLAN}}}
        & \multirow{3}{*}{$\mathcal{A}_{tc}$ ($\uparrow$)}
        & IP & $76.6$  & $75.7$ & $84.0$ & $86.0$ & $77.2$ & $62.3$ & N/A \\
        & & DP & $89.2$  & $97.1$ & $95.2$ & $91.5$ & $88.1$ & $86.2$ & N/A \\
        & & HF & $100.0$ & $100.0$ & $100.0$ & $100.0$ & $100.0$ & $100.0$ & N/A \\
        \cmidrule(lr){2-10}
        & \multirow{3}{*}{$\delta$ ($\downarrow$)}
        & IP & $0.025$ & $0.030$ & $0.022$ & $0.023$ & $0.026$ & $0.026$ & N/A \\
        & & DP & $0.059$ & $0.072$ & $0.053$ & $0.053$ & $0.059$ & $0.067$ & N/A \\
        & & HF & $0.095$ & $0.074$ & $0.079$ & $0.092$ & $0.096$ & $0.103$ & N/A \\
        \cmidrule(lr){2-10}
        & \multirow{3}{*}{$\varepsilon$ (\%, $\downarrow$)}
        & IP & $2.76$  & $8.74$ & $4.20$ & $2.92$ & $2.59$ & $2.33$ & N/A \\
        & & DP & $3.10$  & $6.01$ & $4.29$ & $3.32$ & $2.97$ & $2.65$ & N/A \\
        & & HF & $3.05$  & $5.73$ & $4.14$ & $3.13$ & $2.90$ & $2.82$ & N/A \\
        \cmidrule(lr){2-10}
        & \multirow{3}{*}{$\rho_{\mathrm{out}}$ (\%, $\downarrow$)}
        & IP & $9.22$  & $13.86$ & $10.87$ & $9.12$ & $8.88$ & $9.30$ & N/A \\
        & & DP & $0.05$  & $0.00$ & $0.03$ & $0.04$ & $0.05$ & $0.07$ & N/A \\
        & & HF & $0.00$  & $0.00$ & $0.00$ & $0.00$ & $0.00$ & $0.00$ & N/A \\
        \cmidrule(lr){2-10}
        & \multirow{3}{*}{$\rho_{\mathrm{ovl}}$ (\%, $\downarrow$)}
        & IP & $20.46$ & $20.92$ & $18.64$ & $18.42$ & $20.20$ & $23.81$ & N/A \\
        & & DP & $0.26$  & $0.00$ & $0.15$ & $0.19$ & $0.30$ & $0.31$ & N/A \\
        & & HF & $0.00$  & $0.00$ & $0.00$ & $0.00$ & $0.00$ & $0.00$ & N/A \\
        \midrule
        \multirow{15}{*}{\rotatebox[origin=c]{90}{\makebox[0pt]{Out of Distribution (WMR24)}}}
        & \multirow{3}{*}{$\mathcal{A}_{tc}$ ($\uparrow$)}
        & IP & $2.18$  & $13.4$ & $1.9$ & $3.5$ & $0.0$ & $0.0$ & $0.0$ \\
        & & DP & $83.2$  & $92.0$ & $86.5$ & $82.5$ & $79.7$ & $74.7$ & N/A \\
        & & HF & $100.0$ & $100.0$ & $100.0$ & $100.0$ & $100.0$ & $100.0$ & $100.0$ \\
        \cmidrule(lr){2-10}
        & \multirow{3}{*}{$\delta$ ($\downarrow$)}
        & IP & $0.025$ & $0.022$ & $0.030$ & $0.025$ & $0.026$ & $0.026$ & $0.024$ \\
        & & DP & $0.104$ & $0.095$ & $0.104$ & $0.107$ & $0.104$ & $0.109$ & N/A \\
        & & HF & $0.090$ & $0.065$ & $0.082$ & $0.086$ & $0.085$ & $0.101$ & $0.098$ \\
        \cmidrule(lr){2-10}
        & \multirow{3}{*}{$\varepsilon$ (\%, $\downarrow$)}
        & IP & $14.76$ & $24.53$ & $19.88$ & $16.96$ & $14.90$ & $13.04$ & $9.47$ \\
        & & DP & $8.63$  & $11.41$ & $10.17$ & $7.69$ & $7.50$ & $6.47$ & N/A \\
        & & HF & $6.27$  & $10.57$ & $7.95$ & $6.96$ & $6.55$ & $5.23$ & $4.32$ \\
        \cmidrule(lr){2-10}
        & \multirow{3}{*}{$\rho_{\mathrm{out}}$ (\%, $\downarrow$)}
        & IP & $13.51$ & $9.62$ & $16.81$ & $12.84$ & $12.72$ & $14.64$ & $13.63$ \\
        & & DP & $0.22$  & $0.15$ & $0.42$ & $0.26$ & $0.06$ & $0.17$ & N/A \\
        & & HF & $0.00$  & $0.00$ & $0.00$ & $0.00$ & $0.00$ & $0.00$ & $0.00$ \\
        \cmidrule(lr){2-10}
        & \multirow{3}{*}{$\rho_{\mathrm{ovl}}$ (\%, $\downarrow$)}
        & IP & $16.40$ & $7.72$ & $14.15$ & $14.78$ & $14.65$ & $15.16$ & $21.43$ \\
        & & DP & $3.23$  & $2.04$ & $3.62$ & $3.22$ & $3.56$ & $3.46$ & N/A \\
        & & HF & $0.00$  & $0.00$ & $0.00$ & $0.00$ & $0.00$ & $0.00$ & $0.00$ \\
        \bottomrule
    \end{tabular}
    \label{tab:per_bin_comparison_boundary}
\end{table*}

The per-bin numbers in Table~\ref{tab:per_bin_comparison_boundary} confirm that the aggregate ranking on the boundary-constrained metrics holds across room-count complexity rather than averaging out. On RPLAN, iPLAN's nominal $\delta \in [0.022, 0.030]$ and $\varepsilon \in [2.33\%, 8.74\%]$ are flat across bins but its tiling errors $\rho_{\mathrm{out}}$ and $\rho_{\mathrm{ovl}}$ stay in the $9\!-\!14\%$ and $18\!-\!24\%$ ranges, respectively, so the geometric advantage of iPLAN is uniformly accompanied by per-bin tiling violations rather than concentrated in any one complexity regime; DP holds $\rho_{\mathrm{out}} \leq 0.07\%$ and $\rho_{\mathrm{ovl}} \leq 0.31\%$ across all bins while $\mathcal{A}_{tc}$ degrades only mildly from $97.1\%$ at $\leq\!4$ rooms to $86.2\%$ at $8$ rooms; and HF holds $\mathcal{A}_{tc} = 100.0\%$ and $\rho_{\mathrm{out}}=\rho_{\mathrm{ovl}}=0.00\%$ on every bin (the latter by construction), with $\delta$ rising mildly from $0.074$ at $\leq\!4$ rooms to $0.103$ at $8$ rooms and $\varepsilon$ \emph{decreasing} from $5.73\%$ to $2.82\%$ over the same range as more rooms make per-type area proportions easier to recover. Out of distribution, the gap widens: iPLAN's $\mathcal{A}_{tc}$ collapses from $13.4\%$ at $\leq\!4$ rooms to $0.0\%$ from the $7$-room bin onward, and its $\rho_{\mathrm{ovl}}$ climbs to $21.43\%$ in the $\geq\!9$ bin where iPLAN no longer recovers a single complete room set; DP retains a graceful degradation pattern ($\mathcal{A}_{tc} \in [74.7\%, 92.0\%]$, $\rho_{\mathrm{ovl}} \in [2.04\%, 3.62\%]$) but is undefined for $\geq\!9$ rooms because no plans of that size are present in its evaluation split; HF is the only method that maintains $\mathcal{A}_{tc} = 100.0\%$ and $\rho_{\mathrm{out}} = \rho_{\mathrm{ovl}} = 0.00\%$ in every WMR24 bin including the largest one, with $\delta$ rising from $0.065$ at $\leq\!4$ rooms to $0.098$ at $\geq\!9$ rooms while $\varepsilon$ again \emph{decreases} from $10.57\%$ to $4.32\%$ as the room count grows.

\begin{table*}[h]
    \centering
    \footnotesize
    \caption{Per-bin comparison on \textbf{RPLAN} and \textbf{WMR24} for HypergraphFormer trained on progressively smaller subsets, reporting GED \eqref{eq:ged} and GED accuracy $\mathcal{A}$ \eqref{eq:accuracy} across room-count bins. All rows report our method (HypergraphFormer); rows differ only in training-set size. The \emph{Full Dataset} rows repeat per-bin cells from the \emph{Optimize $\delta$ and $\varepsilon$} variant in Tables~\ref{tab:per_bin_hf_ablation_rplan} (RPLAN, $12{,}002$ samples) and~\ref{tab:per_bin_hf_ablation_wmr24} (WMR24, $1{,}111$ samples); the remaining rows correspond to training sizes $1{,}000$, $5{,}000$, $10{,}000$, and $25{,}000$. Cells marked \texttt{--} indicate that no bin-wise statistics are available.}
    \renewcommand{\arraystretch}{1.05}
    \begin{tabular}{l l l c c c c c c}
        \toprule
        Dataset & Metric & Variant & $\leq 4$ & 5 & 6 & 7 & 8 & $9 \leq$ \\
        \midrule
        \multirow{10}{*}{\rotatebox[origin=c]{90}{\makebox[0pt]{RPLAN}}}
        & \multirow{5}{*}{GED ($\downarrow$)}
        & Full Dataset & $1.50$ & $1.45$ & $1.58$ & $1.63$ & $1.71$ & N/A \\
        & & 1{,}000 samples & $4.73$ & $4.51$ & $4.60$ & $4.73$ & $4.79$ & N/A \\
        & & 5{,}000 samples & $3.69$ & $3.41$ & $3.55$ & $3.71$ & $3.85$ & N/A \\
        & & 10{,}000 samples & $2.31$ & $2.68$ & $2.89$ & $3.04$ & $3.15$ & N/A \\
        & & 25{,}000 samples & $1.74$ & $1.97$ & $1.92$ & $1.99$ & $2.04$ & N/A \\
        \cmidrule(lr){2-9}
        & \multirow{5}{*}{$\mathcal{A}$ ($\uparrow$)}
        & Full Dataset & $46.3$ & $45.8$ & $41.7$ & $40.6$ & $38.8$ & N/A \\
        & & 1{,}000 samples & $4.0$ & $5.3$ & $4.5$ & $4.2$ & $4.0$ & N/A \\
        & & 5{,}000 samples & $14.3$ & $13.4$ & $10.6$ & $9.0$ & $8.8$ & N/A \\
        & & 10{,}000 samples & $28.6$ & $21.1$ & $18.0$ & $15.6$ & $14.9$ & N/A \\
        & & 25{,}000 samples & $42.9$ & $34.6$ & $34.4$ & $32.1$ & $32.3$ & N/A \\
        \midrule
        \multirow{10}{*}{\rotatebox[origin=c]{90}{\makebox[0pt]{WMR24}}}
        & \multirow{5}{*}{GED ($\downarrow$)}
        & Full Dataset & $1.44$ & $1.60$ & $1.60$ & $1.68$ & $1.91$ & $1.86$ \\
        & & 1{,}000 samples & $3.68$ & $3.92$ & $4.01$ & $4.12$ & $3.83$ & $4.21$ \\
        & & 5{,}000 samples & $3.02$ & $3.06$ & $3.18$ & $3.26$ & $3.78$ & $3.25$ \\
        & & 10{,}000 samples & $2.21$ & $2.85$ & $2.97$ & $2.36$ & $3.27$ & $2.88$ \\
        & & 25{,}000 samples & $1.98$ & $1.76$ & $1.94$ & $2.44$ & $1.88$ & $2.26$ \\
        \cmidrule(lr){2-9}
        & \multirow{5}{*}{$\mathcal{A}$ ($\uparrow$)}
        & Full Dataset & $56.0$ & $52.2$ & $53.5$ & $51.3$ & $50.1$ & $49.8$ \\
        & & 1{,}000 samples & $17.2$ & $9.3$ & $11.3$ & $13.3$ & $18.2$ & $10.6$ \\
        & & 5{,}000 samples & $21.2$ & $18.8$ & $24.0$ & $21.7$ & $14.6$ & $24.6$ \\
        & & 10{,}000 samples & $32.6$ & $26.3$ & $26.0$ & $32.3$ & $22.3$ & $25.7$ \\
        & & 25{,}000 samples & $39.8$ & $50.0$ & $44.4$ & $40.7$ & $46.8$ & $41.4$ \\
        \bottomrule
    \end{tabular}
    \label{tab:per_bin_rplan_data_scaling}
\end{table*}

\section{Converting RPLAN to the hypergraph format}
\label{app:rplan_to_hypergraph}

\begin{figure*}[t]
    \centering
    \includegraphics[width=\linewidth]{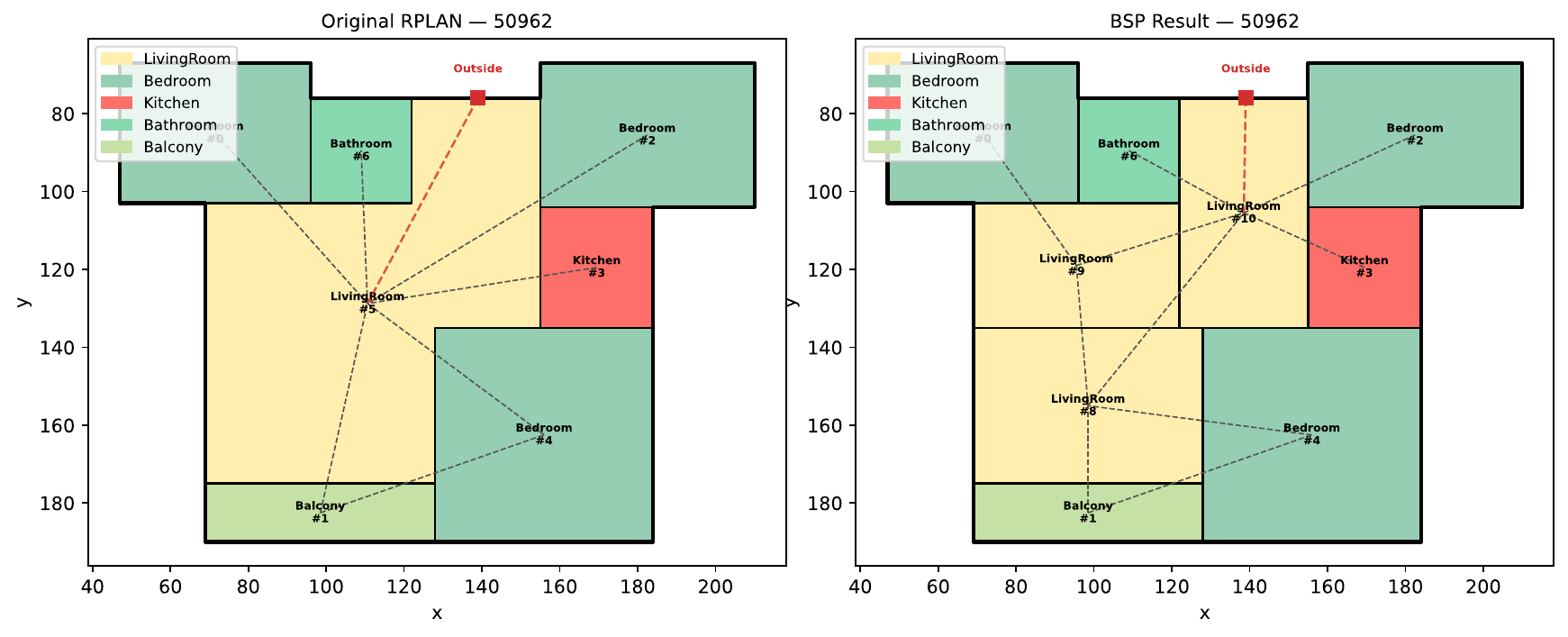}
    \\[4pt]
    \includegraphics[width=\linewidth]{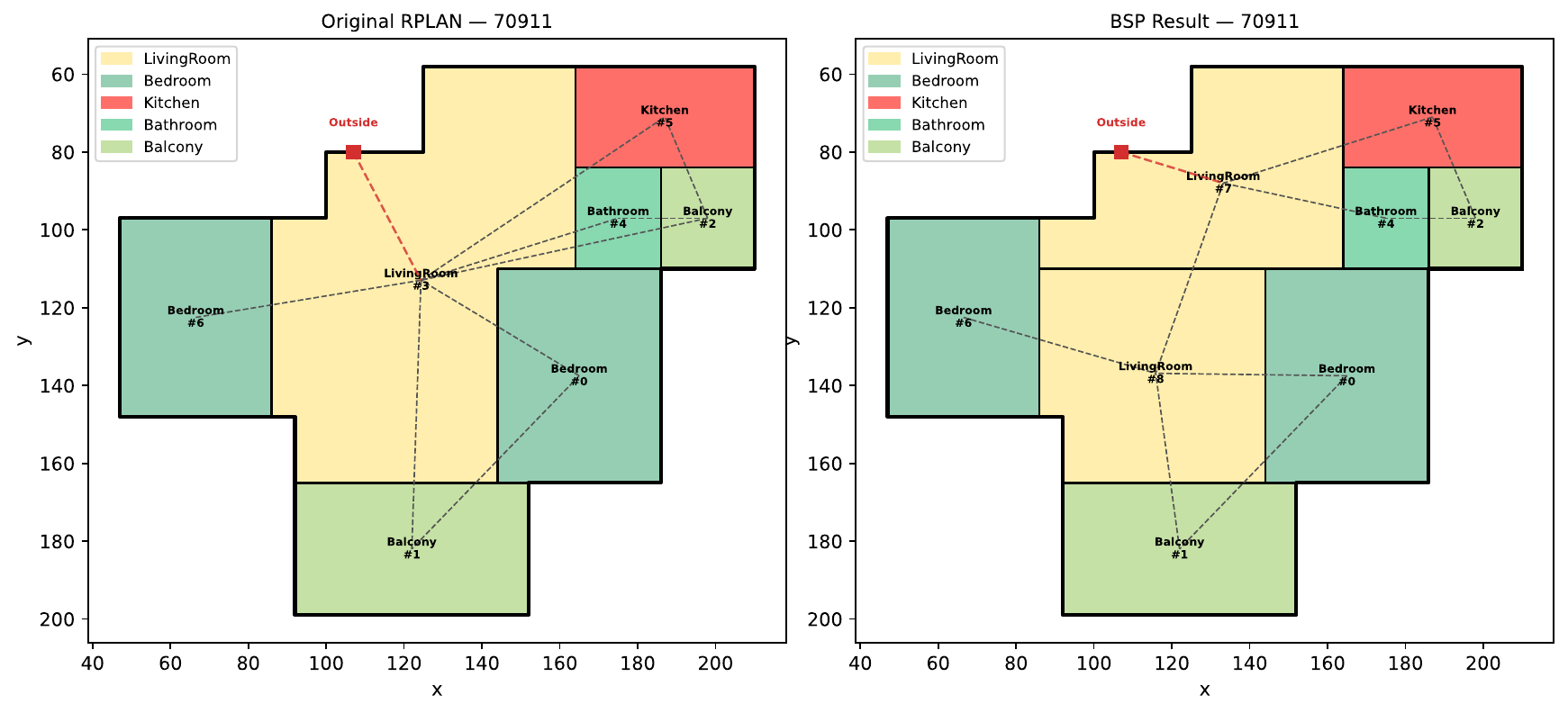}
    \caption{RPLAN-to-hypergraph conversion on two test samples (\texttt{50962}, top; \texttt{70911}, bottom). Each row shows the original RPLAN plan with its door-based access graph and entrance (\textit{left}) and the BSP-derived layout with leaf-level rooms, where the LivingRoom has been subdivided into geometrically adjacent sibling fragments and every original access edge has been re-routed to a unique leaf pair (\textit{right}). The two panels share identical visual styling. The dashed grey segments are access edges and the red square marks the \texttt{Outside} (entrance) node.}
    \label{fig:rplan_bsp_examples}
\end{figure*}

In addition to our WMR24 dataset, we also evaluate on the widely used RPLAN benchmark~\cite{RPLAN}. RPLAN distributes each plan as a list of axis-aligned room rectangles, an outer boundary polygon, an entrance rectangle, and a door-based \emph{access adjacency} list. To make these plans usable by HypergraphFormer, we convert every sample into the same textual hypergraph (BSP-tree) format that the model emits during inference. This appendix summarizes the conversion algorithm and illustrates its output on two representative samples (Fig.~\ref{fig:rplan_bsp_examples}).

\paragraph{Input parsing.}
For each RPLAN sample we read the per-room polygons from the \texttt{r\_boundary} field, the outer boundary, the door-based connectivity from \texttt{access\_adjacencies}, and the entrance rectangle from \texttt{entrance\_expand}. Rooms with degenerate or near-zero-area polygons are discarded, and the outer boundary is recomputed as the union of all room polygons (using the largest connected component when the union is multi-part) so that the boundary used for splitting is exactly consistent with the rooms it contains.

\paragraph{Recursive binary space partitioning.}
The hypergraph format represents a layout as a binary BSP tree whose leaves are the rooms and whose internal nodes are axis-aligned splits of the parent region. Given the rooms and outer boundary, we build this tree top-down. At each node, we enumerate every candidate horizontal and vertical line whose position coincides with at least one room vertex (so that the search is finite and aligned with the actual layout grid), and for each candidate we classify the rooms into a low side, a high side, and a set of straddling rooms that the line would cut. We score each valid candidate using a tiered objective:
\begin{enumerate}
    \item[(0)] a \emph{clean} split that does not cut any room;
    \item[(1)] a split that cuts only \texttt{LivingRoom}(s), with both line endpoints on the outer boundary;
    \item[(2)] a split that cuts only \texttt{LivingRoom}(s) but with at least one endpoint interior;
    \item[(3)] a split that cuts a non-\texttt{LivingRoom}, with both endpoints on the boundary;
    \item[(4)] anything else (last resort).
\end{enumerate}
Within a tier, candidates are further ranked by how few rooms they cut, how evenly each cut room is divided, and how balanced the overall area split is. Cutting the LivingRoom is privileged because in real plans the LivingRoom is typically the circulation hub that connects to all other rooms and naturally hosts the principal partition lines, so subdividing it yields cleaner sub-trees than cutting an enclosed bedroom or bathroom. To avoid greedy traps, the search optionally evaluates each candidate one level ahead: for every otherwise-valid split it simulates the best greedy split on each of the two resulting sub-regions and adds a small fraction of those scores to the parent's score, biasing the choice towards splits that lead to cleaner downstream partitions. As an alternative we also experimented with an evolutionary search that samples among same-tier candidates with a population-based procedure optimizing a global tree-quality objective (per-leaf aspect ratio, total cuts, and tree balance); we use the default look-ahead-of-one greedy variant for the experiments in this paper as we found it sufficient on RPLAN.

\paragraph{Splitting non-convex rooms and discarding spurious fragments.}
When a chosen split line passes through one or more straddling rooms, each such room is intersected with the two half-planes defined by the line. Because RPLAN rooms are not always rectangular, this intersection can produce more than one polygon on a single side; we keep all sub-polygons whose area exceeds a small floor of $10^{-6}$ as separate fragments, and assign each a fresh globally unique room ID via a shared counter that is threaded through the recursion. For non-LivingRoom rooms we additionally discard any fragment whose area is below $5\%$ of the original room's area: such tiny slivers are almost always artifacts of a split line that grazes the room corner rather than a meaningful subdivision.

\paragraph{Access-edge propagation.}
Every original access edge $(u,v)$ from the door graph must be re-attached to leaf-level rooms in the final tree. Because $u$ or $v$ may have been subdivided into multiple fragments at one or more levels of the recursion, we resolve each original edge to the unique pair of descendant fragments $(u^\star, v^\star)$ whose polygons share the longest non-trivial wall (we require shared length $> 0.5$ to avoid spurious point-only contacts that occasionally appear in the raw RPLAN labels). In addition, whenever a single original room is split into multiple sibling fragments we connect any pair of those fragments that shares a wall, so that the original room's interior remains traversable in the final access graph. Finally, the entrance rectangle is matched to the leaf room whose polygon overlaps it most (falling back to the nearest centroid when there is no overlap), and a synthetic edge from a designated \texttt{Outside} node to that leaf is added to encode the front door.

\paragraph{Hypergraph JSON export.}
With the BSP tree, leaf polygons, access edges, and \texttt{Outside} edge in hand, the conversion writes one JSON entry per plan in exactly the schema HypergraphFormer consumes for our WMR24 dataset. Each tree node is given a unique path-based name (\texttt{root}, \texttt{rootL}, \texttt{rootLR}, \dots) so that internal nodes and leaves can be referenced unambiguously in the textual representation. Each leaf carries its area, room category (mapped to one of \texttt{living}/\texttt{bed}/\texttt{kitchen}/\texttt{bath}/\texttt{balcony}/\texttt{storage}), the path names of its access neighbors, and the split angle inherited from its parent. Each internal node carries its split angle and its two children. To match the convention used by the renderer that reconstructs geometry from the BSP tree, children of horizontal splits are reordered so that the first child is the one on the lower side. The boundary polygon is exported in two redundant forms (a list of corner points and a list of facade segments, both with $z=0$ for compatibility with our 3-D-capable schema), together with bookkeeping metadata such as the source database, the bedroom and bathroom counts, and the total area.

\paragraph{Examples.}
Fig.~\ref{fig:rplan_bsp_examples} visualizes the conversion on two RPLAN test samples. In both rows the left subplot shows the original RPLAN rooms with their door-based access graph and the entrance marker, and the right subplot shows the BSP-derived layout with leaf-level rooms (note the LivingRoom subdivided into multiple sibling fragments that remain mutually adjacent in the access graph). Sample \texttt{50962} (top) is a fairly regular plan where two horizontal cuts at the top suffice to peel off the bedrooms, after which the LivingRoom is split once vertically; sample \texttt{70911} (bottom) is more irregular and requires several nested splits, with the LivingRoom subdivided three times to give every other room its own clean sibling region. In both cases every original RPLAN room is preserved (possibly as multiple sibling fragments) and every door-based access edge is resolved to a leaf-level pair, yielding a complete and self-contained BSP-tree description of the layout that HypergraphFormer can be trained to generate and edit.

\section{Ablation study on LoRA configuration}
\label{app:lora_ablation}
We conduct an ablation study on the configuration of Low-Rank Adaptation (LoRA) to assess the impact of different parameter settings on model performance. Specifically, we vary the LoRA rank $r$ and scaling factor $\alpha$, while keeping all other training hyperparameters fixed. For each configuration, the model is trained using supervised fine-tuning and evaluated on a held-out validation set. Model selection is performed based on the next-token prediction loss on the validation set, and we select the final model corresponding to the configuration that achieves the lowest validation loss. This ablation allows us to quantify the trade-off between model capacity, parameter efficiency, and task performance.

\begin{table}[t]
    \centering
    \small
    \caption{Ablation study of LoRA configurations. Model selection is based on the next-token prediction loss on the validation set.}
    \begin{tabular}{c c c}
        \toprule
        $r$ (Rank) & $\alpha$ (Scaling) & Validation Loss ($\downarrow$) \\
        \midrule
        32 & 64  &  0.2619\\
        64  & 128  & \textbf{0.2587} \\
        128 & 256 & 0.2599\\
        \bottomrule
    \end{tabular}
    \label{tab:lora_ablation}
\end{table}

The results in Table~\ref{tab:lora_ablation} indicate that increasing the LoRA rank and scaling factor initially leads to improved validation performance, with the configuration $r=64$, $\alpha = 128$ achieving the lowest next-token prediction loss. This suggests that a moderate increase in adaptation capacity is beneficial for the task, likely enabling the model to capture task-specific patterns more effectively than lower-rank configurations. However, further increasing the rank to $r=128$, with $\alpha=256$, does not yield additional gains and instead slightly degrades validation loss. This behavior points to diminishing returns from higher-capacity LoRA configurations, where additional parameters may not translate into improved generalization. Overall, the results highlight a trade-off between parameter efficiency and performance, with an intermediate LoRA configuration offering the best balance for this setting.

\section{The WMR24 dataset}
\label{app:dataset_wmr24}

\begin{figure}[h]
    \centering
    \includegraphics[width=\linewidth]{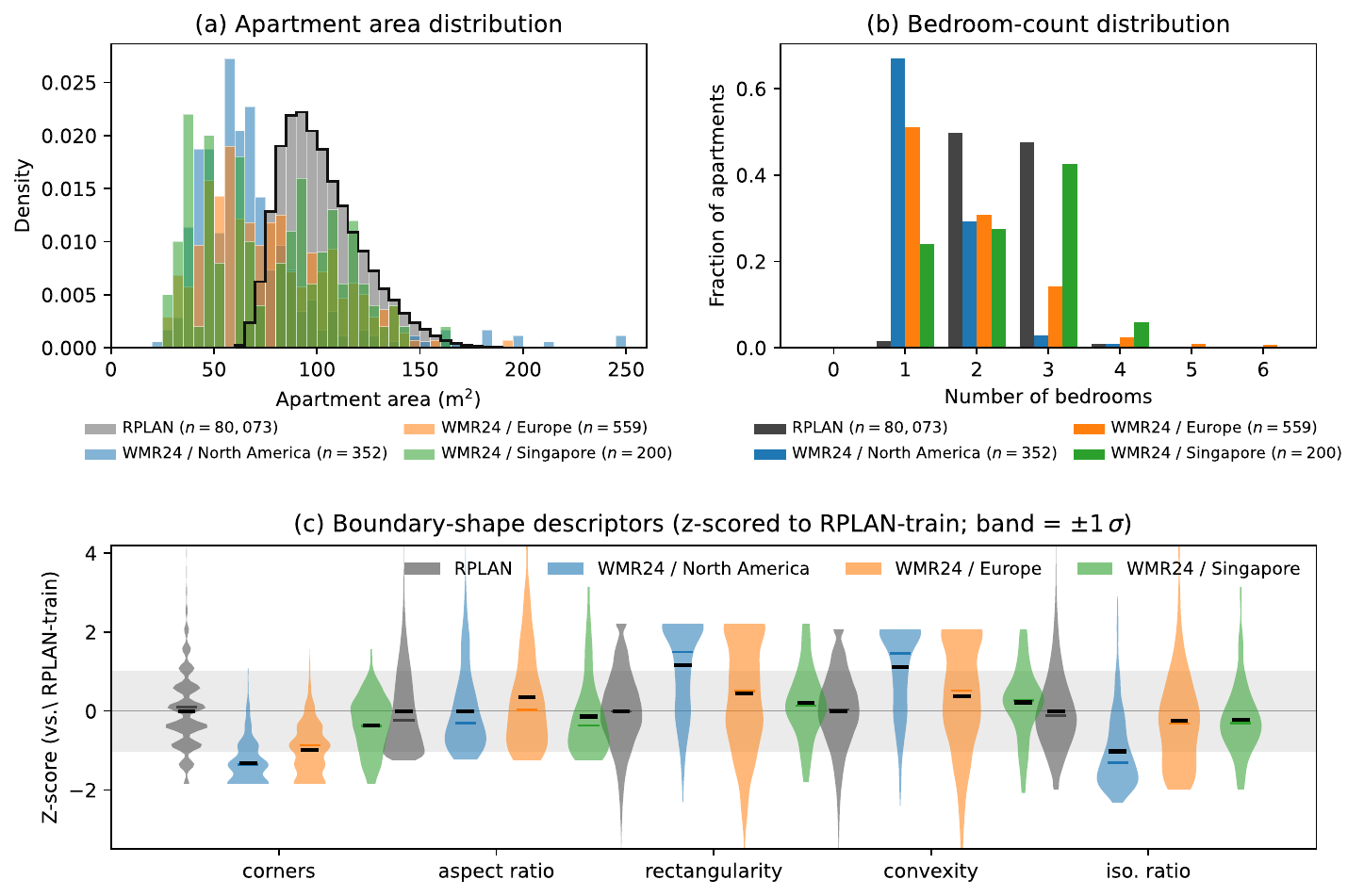}
    \caption{Distribution of (a) apartment area, (b) number of bedrooms, and (c) five boundary-shape descriptors across RPLAN (gray, $n = 80{,}073$) and the three WMR24 sub-regions used as out-of-distribution test data. In (a) the dark gray outline is the RPLAN profile re-drawn on top of the WMR24 layers so the in-distribution silhouette remains visible where WMR24 bars otherwise occlude it. In (c) we plot, per group, the distribution of: corner count (number of polygon vertices on the apartment boundary after merging collinear vertices); aspect ratio of the minimum-area oriented bounding box; rectangularity (plan area divided by oriented-bounding-box area, in $(0, 1]$, with $1$ a perfect rectangle); convexity (plan area divided by convex-hull area); and isoperimetric ratio ($P / (2\sqrt{\pi A})$, $\geq 1$). All five descriptors are z-scored using RPLAN-train statistics, so $0$ marks the RPLAN-train mean and the gray band marks $\pm 1$ RPLAN-train standard deviation; black tick marks indicate group means and the inner horizontal lines indicate group medians. RPLAN is sharply concentrated near $100$ m$^2$ with $2$--$3$ bedrooms and complex axis-aligned L/T-shaped boundaries, whereas WMR24 has a substantially wider area range, a much heavier tail of small ($<60$ m$^2$) units, a bedroom distribution centred on $1$--$2$ bedrooms with non-trivial mass at studios, and -- particularly for the North American and European subsets -- a clear shift toward simpler, more rectangular boundaries with fewer corners. The three axes together illustrate the distribution shift that the WMR24 evaluations in the main paper measure.}
    \label{fig:dataset_comp}
\end{figure}

\emph{WMR24}~\cite{hypergraph} is a curated collection of $1{,}111$ architect-designed and real-world floor plans assembled from eleven sub-collections spanning North America, Europe, and Singapore, ranging from studios to six-bedroom apartments. We use it strictly as an out-of-distribution \emph{test} set; HypergraphFormer and the rasterized/vectorized baselines we compare against are trained only on RPLAN~\cite{RPLAN}, which we access through the splits provided by~\cite{DiffPlanner} ($56{,}053$ training, $12{,}018$ validation, and $12{,}002$ test plans, with no overlap to WMR24). Per-region statistics for WMR24 and aggregate statistics for RPLAN are reported in Table~\ref{tab:dataset_stats}, and the corresponding distributions are visualized in Fig.~\ref{fig:dataset_comp}; together they show that WMR24 differs from RPLAN on the three axes that matter most for boundary-constrained floor plan generation: apartment size, bedroom count, and the geometry of the apartment boundary itself.

Despite RPLAN's scale ($\sim$$80{,}000$ plans), its diversity is narrow. Every RPLAN plan lies in a tight band of $63.2$--$201.4$ m$^2$ (mean $102.0 \pm 19.8$ m$^2$) and has either 2 or 3 bedrooms ($98\%$ of the dataset, mean $2.48 \pm 0.55$). WMR24 is roughly $70\times$ smaller but markedly broader: areas range from $21.5$ to $246.4$ m$^2$ (mean $74.0 \pm 31.1$ m$^2$, more than $1.5\times$ the RPLAN spread), and the bedroom distribution covers $0$--$6$ bedrooms with a mean of $1.72 \pm 0.89$. The overlap is partial: WMR24 contains an entire regime of small, low-bedroom apartments (studios and one-bedroom units, particularly in North America) that RPLAN essentially does not represent, and it also includes very large ($>200$ m$^2$) plans that lie at the extreme tail of RPLAN.

The boundary geometry shows the same pattern. RPLAN boundaries are exclusively axis-aligned and concentrate near complex L/T-shapes with $8$--$14$ corners; WMR24 boundaries -- particularly the North American and European subsets -- are shifted toward simpler near-rectangular shapes (Fig.~\ref{fig:dataset_comp}c; mean $6.2$ corners and rectangularity $0.91$ for North America vs.\ $11.6$ and $0.80$ for RPLAN; Cohen's $d \approx 1.1$--$1.3$ on four of five shape descriptors), and contain a small but non-trivial fraction ($\approx$$4\%$ North America, $\approx$$2\%$ Europe) of plans with non-axis-aligned walls that RPLAN cannot represent at all. The Singapore subset, by contrast, follows conventions closer to RPLAN's grid-housing tradition and shows much smaller shifts on every descriptor ($d < 0.4$); the boundary-shape OOD signal in WMR24 is concentrated in the North American and European collections. Although only $\sim$$2\%$ of WMR24 plans fall outside RPLAN's $99\%$ Mahalanobis ellipsoid in shape space, fewer than half of North American plans lie inside RPLAN's central $90\%$ box on all five shape descriptors jointly, indicating a substantial mean shift in the boundary-shape distribution rather than the appearance of a wholly new shape regime.

This combination of distribution shift in size, bedroom count, boundary geometry, and underlying design conventions makes WMR24 a strict OOD evaluation for any model trained on RPLAN, and it is the basis for the OOD generalization claims reported in the main paper. We refer the reader to~\cite{WEBER2022} for a broader discussion of why more diverse, professionally curated benchmarks are needed for floor plan generation.

\begin{table}[h]
    \centering
    \footnotesize
    \renewcommand{\arraystretch}{1.05}
    \setlength{\tabcolsep}{5pt}
    \begin{tabular}{@{}l c c c c c c@{}}
        \toprule
        \textbf{Subset}
        & \textbf{Plans}
        & \textbf{Area (m$^2$)}
        & \textbf{Area range}
        & \textbf{Bedrooms}
        & \textbf{Corners}
        & \textbf{Rect.} \\
        \midrule
        \multicolumn{7}{@{}l}{\emph{WMR24 (out-of-distribution test set)}} \\
        \quad North America & $352$  & $68.2 \pm 31.6$ & $21.5$--$246.4$ & $1.38 \pm 0.59$ & $6.2 \pm 2.3$ & $0.91 \pm 0.10$ \\
        \quad Europe        & $559$  & $76.3 \pm 29.7$ & $27.4$--$190.8$ & $1.74 \pm 0.94$ & $7.5 \pm 2.7$ & $0.84 \pm 0.13$ \\
        \quad Singapore     & $200$  & $77.8 \pm 32.8$ & $26.0$--$162.0$ & $2.31 \pm 0.90$ & $10.1 \pm 2.9$ & $0.82 \pm 0.08$ \\
        \quad \textbf{WMR24 total} & $\mathbf{1{,}111}$ & $\mathbf{74.0 \pm 31.1}$ & $\mathbf{21.5}$--$\mathbf{246.4}$ & $\mathbf{1.72 \pm 0.89}$ & $\mathbf{7.5 \pm 2.9}$ & $\mathbf{0.86 \pm 0.12}$ \\
        \midrule
        \multicolumn{7}{@{}l}{\emph{RPLAN (training source; train/val/test $= 56{,}053/12{,}018/12{,}002$)}} \\
        \quad RPLAN total   & $80{,}073$ & $102.0 \pm 19.8$ & $63.2$--$201.4$ & $2.48 \pm 0.55$ & $11.6 \pm 4.1$ & $0.80 \pm 0.09$ \\
        \bottomrule
    \end{tabular}
    \caption{Dataset statistics. Areas are reported in m$^2$ as mean $\pm$ standard deviation, with min--max in the third column. WMR24 areas are read directly from the dataset's metadata, which stores them in m$^2$. RPLAN areas are obtained by summing per-room pixel areas from each plan and converting using the standard $256$-px-per-$18$-m grid convention (one pixel side $= 18/256 \approx 0.070$ m, one pixel area $\approx 4.94 \times 10^{-3}$ m$^2$). Bedrooms are counted as rooms with the canonical RPLAN bedroom category; for WMR24, the dataset's per-sample \texttt{bedrooms} field is used directly. \emph{Corners} is the number of polygon vertices on the apartment boundary after merging collinear vertices (so door/window markers along straight walls are not double-counted). \emph{Rect.}\ (rectangularity) is the plan area divided by the area of its minimum-area oriented bounding box, in $(0, 1]$, so that $1.0$ corresponds to a perfect rectangle. The full set of five boundary-shape descriptors is shown in Fig.~\ref{fig:dataset_comp}c.}
    \label{tab:dataset_stats}
\end{table}

\section{Limitations and future work}
\label{app:future_work}
\label{app:hypergraph_limitations}

The hypergraph is not an explicit data format but rather a sequence of geometric operations that, when applied, recover the internal layout of a floor plan. Compared to image- or vector-based representations, this implicit formulation has a few limitations that we acknowledge here, together with the directions for future work that they motivate.

\paragraph{Implicit geometry recovery.}
Because rooms are represented through a BSP tree, they are easy to compare in graph form, but recovering the actual room geometry requires running the subdivision algorithm. If subdivisions are precomputed for certain boundaries and the boundaries are subsequently changed, the geometry can become inconsistent with the cached splits unless the tree is re-evaluated on the new boundary. Future work could explore incremental re-evaluation strategies, or a hybrid representation that caches geometry alongside the BSP tree and keeps the two synchronized under boundary edits.

\paragraph{Beyond rectilinear partitions.}
The current BSP formulation assumes axis-aligned or angled straight-line splits, which already captures a wide range of real apartments (including the non-Manhattan layouts shown in our experiments) but excludes curved walls and other non-polygonal partitions. Extending the representation to richer split primitives, while preserving the boundary-independent property that makes the hypergraph easy to learn, is an interesting avenue for future work. For example, introducing notations to represent rooms as groups of partitions may further adapt the hypergraph format to complex and non-convex room layouts, further improving its similarity to ground-truth room contours. Accounting for wall thickness may allow to expand it to datasets like MSD~\cite{van2024msd, engelenburgGKN_2025}.

\paragraph{Scaling to multi-unit and full-building layouts.}
We focus on single-apartment floor plans. Scaling the same representation to multi-unit floors or whole buildings would require composing hypergraphs across units (corridors, shafts, shared services) and integrating program-level constraints. We see this as a promising direction, particularly in combination with the tool-call editing interface, which already exposes the BSP tree and access graph as first-class objects an LLM can manipulate.

\paragraph{Learning to edit with hypergraph representation in mind.}
Even though we have demonstrated that generated hypergraphs can be easily augmented with simple procedural edits, we envision further opportunities to modify existing hypergraphs against specific objectives and with better awareness of the representation via reinforcement learning or more complex optimization techniques. Beyond repairing hypergraphs, we envision more complex optimization metrics, such as wind, daylight performance, or occupancy.


\end{document}